\documentclass[preprint,authoryear,11pt]{elsarticle}

\usepackage{graphicx}

\usepackage{amssymb}

\usepackage[top=2.2cm,left=2.1cm,right=2.1cm,bottom=2.1cm]{geometry}

\usepackage{enumitem} 
  \setlist{itemsep=1ex plus0.2ex, leftmargin=*, align=left}

\makeatletter
\newcommand{\labitem}[2]{%
\def\@itemlabel{\textbf{#1}}
\item
\def\@currentlabel{#1}\label{#2}}
\makeatother

\makeatletter
\newcommand{\headingitem}[1]{%
\vspace{0.3cm}
\def\@itemlabel{\textbf{#1}}
\item
\def\@currentlabel{#1}
\addtocounter{enumi}{-1}
}
\makeatother

\usepackage{csvsimple}
\usepackage{multirow}
\usepackage{lipsum}
\usepackage{afterpage}

\usepackage{etoolbox}
\robustify\bfseries

\usepackage{eurosym}
\usepackage{siunitx}

    \DeclareSIUnit\eur{\officialeuro}
    \DeclareSIUnit\M{M}
    \DeclareSIUnit\k{k}

  \def\sym#1{\ifmmode^{#1}\else\(^{#1}\)\fi}

\usepackage{setspace}
\usepackage{csquotes}
\usepackage{amsmath}
\usepackage{bm}

\usepackage{xspace}
	\newcommand\ie{i.\,e.\xspace}
	\newcommand\eg{e.\,g.\xspace}

	\newcommand\cf{cf.\xspace}
	
	\newcommand\cp{cp.\xspace}

\usepackage[amsmath,hyperref,framed]{ntheorem} 
  \theoremstyle{plain}
     
  \theoremstyle{nonumberplain}
    \theoremseparator{.}
    \theoremheaderfont{\bfseries}
    \theorembodyfont{\normalfont}
    \theoremsymbol{$\blacksquare$}
      \RequirePackage{amssymb}

      \makeatletter
    \let\copy@theorem@headerfont=\theorem@headerfont
    \newcommand{\my@theorem@headerfont}{%
        \boldmath\copy@theorem@headerfont\unboldmath
      }
    \let\theorem@headerfont=\my@theorem@headerfont
      \makeatother

\theoremstyle{nonumberplain}
\theoremseparator{.}
\newcommand{\argmin}{\operatornamewithlimits{arg \, min}}

\usepackage[%
  sort&compress
]{cleveref}
\usepackage{url}

\usepackage{isomath}

\newcommand{\mathup}[1]{\mathrm{#1}}

\newcommand{\e}[1]{\mathup{e}^{#1}}

  \newcommand{\norm}[1]{\left\lVert#1\right\rVert}

\usepackage{algorithm}
\usepackage{algpseudocode}
\algnewcommand{\LineComment}[1]{\State \(\triangleright\) #1}
\algdef{SE}[DOWHILE]{Do}{doWhile}{\algorithmicdo}[1]{\algorithmicwhile\ #1}%



\usepackage[usenames,dvipsnames]{xcolor}

\usepackage{colortbl}
\usepackage{tabularx}
\usepackage{booktabs}
\usepackage{collcell}

\usepackage{ragged2e}
  
\newcommand{\PreserveBackslash}[1]{\let\temp=\\#1\let\\=\temp}
\newcolumntype{v}[1]{>{\PreserveBackslash\RaggedRight\hspace{0pt}}p{#1}}

  \usepackage{adjustbox}
\newcolumntype{Q}[2]{%
    >{\adjustbox{angle=#1,lap=\width-(#2)}\bgroup}%
    l%
    <{\egroup}%
}

\newcommand{\mcellt}[2][c]{%
  \begin{tabular}[t]{@{}#1@{}}#2\end{tabular}}

\usepackage[
    colorinlistoftodos,
    textsize=footnotesize,
        ]{todonotes}

\makeatletter
    \renewcommand{\fps@figure}{htb}         
    \renewcommand{\fps@table}{htb}         
\makeatother 

\usepackage{float}
\usepackage{rotating}

\journal{European Journal of Operational Research}

\begin{document}

\begin{frontmatter}



\title{Deep learning in business analytics and operations research:\\ Models, applications and managerial implications\\[1.2em]
\begin{center}
\normalsize
\emph{Submitted to special issue on\\ \textquote{Business Analytics: Defining the field and identifying a research agenda}}
\end{center}
}


\author[ETH]{Mathias Kraus}
\ead{mathiaskraus@ethz.ch}

\author[ETH]{Stefan Feuerriegel}
\ead{sfeuerriegel@ethz.ch}

\author[Umass]{Asil Oztekin\corref{cor1}}
\ead{asil\_oztekin@uml.edu}

\address[ETH]{ETH Zurich, Weinbergstr. 56/58, 8092 Zurich, Switzerland}
\address[Umass]{Department of Operations \& Information Systems, Manning School of Business, University of Massachusetts Lowell, Lowell, MA 01854 USA}

\cortext[cor1]{Corresponding author. Technical questions should be directed to Mathias Kraus.}

\begin{abstract}
Business analytics refers to methods and practices that create value through data for individuals, firms, and organizations. This field is currently experiencing a radical shift due to the advent of deep learning: deep neural networks promise improvements in prediction performance as compared to models from traditional machine learning. However, our research into the existing body of literature reveals a scarcity of research works utilizing deep learning in our discipline. Accordingly, the objectives of this overview article are as follows: (1)~we review research on deep learning for business analytics from an operational point of view. (2) We motivate why researchers and practitioners from business analytics should utilize deep neural networks and review potential use cases, necessary requirements, and benefits. (3)~We investigate the added value to operations research in different case studies with real data from entrepreneurial undertakings. All such cases demonstrate improvements in operational performance over traditional machine learning and thus direct value gains. (4)~We provide guidelines and implications for researchers, managers and practitioners in operations research who want to advance their capabilities for business analytics with regard to deep learning. (5)~Our computational experiments find that default, out-of-the-box architectures are often suboptimal and thus highlight the value of customized architectures by proposing a novel deep-embedded network.
\end{abstract}

\end{frontmatter}



\section{Introduction}
\label{sec:intro}

What was once described by \citet{Davenport.2007} as \emph{\textquote{firms compet[ing] on analytics}} is now more true than ever. This development is currently propelled by the surge in \textquote{big data}, allowing for an efficient access and analysis of large datasets \citep{Chen.2012, Agarwal.2014, Baesens.2016}. Innovations in business analytics have become not only desirable but a key necessity for the successful performance of firms \citep{Mortenson.2015,Lim.2013,Ranyard.2015}. This holds true for all areas of business operations. Examples include, for instance, supply chain management \citep{Carbonneau.2008}, risk modeling \citep{Lessmann.2015}, commerce \citep{Scholz.2017}, preventive maintenance \citep{Sun.2009}, and manufacturing \citep{Hu.2017}. The core component for successfully competing with business analytics is the underlying predictive model. This represents the unit responsible for making the actual forecasts and its accuracy contributes directly to the overall value creation. 

\begin{figure}[H]
	\centering
	\includegraphics[width=.22\textwidth]{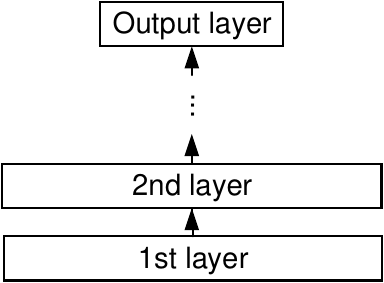}
	\caption{Illustrative deep neural network with four layers. Here the width of a layer corresponds to its dimension, \ie the number of neurons.}
	\label{fig:MLP}
\end{figure}

With recent advances in machine learning, a specific type of predictive model has received great traction lately: \emph{deep learning} \citep{LeCun.2015}. The underlying concept is not specific to machine learning or data-analytics approaches from operations research, as it simply refers to \emph{deep} neural networks. However, what has changed from early experiments with neural networks is the dimension of the networks, which now can easily contain up to hundreds of layers, millions of neurons and complex structures of connections between them \citep[e.\,g.][]{He.2016}. This introduces the unprecedented flexibility to model even highly complex, non-linear relationships between predictor and outcome variables, a quality that has allowed deep neural networks to outperform models from traditional machine learning in a variety of tasks. \Cref{fig:MLP} depicts an illustrative sketch of a deep neural network. It should also be noted that big data \citep{George.2014, George.2016} is now prevalent in firms, yet traditional machine learning (\eg, support vector machines, random forests) can often not be applied to such large datasets, whereas the optimization routines in deep learning scale efficiently.

The expected improvement in prediction performance provided by deep learning has led to a selection of showcases.\footnote{\emph{Harvard Business Review} (2017): \textquote{Deep learning will radically change the ways we interact with technology} \url{https://hbr.org/2017/01/deep-learning-will-radically-change-the-ways-we-interact-with-technology}, accessed February~20, 2018.} For example, computerized personal assistants, such as Apple's Siri, Amazon's Alexa, Google Now or Microsoft's Cortana, now make heavy use of deep neural networks to recognize, understand and answer human questions.\footnote{\emph{The Economist} (2017): \textquote{Technology quarterly: Finding a voice}. \url{http://www.economist.com/technology-quarterly/2017-05-01/language}, accessed February~20, 2018.} In this regard, Microsoft unveiled a speech recognition system in 2016 that is capable of transcribing spoken words almost as accurately as professionally trained humans. In October 2016, Google launched an update to its translation system that utilizes deep learning in order to improve translation accuracy, thereby approaching the performance of humans.\footnote{\emph{Washington Post} (2016): \textquote{Google Translate is getting really, really accurate}. Available via \url{https://www.washingtonpost.com/news/innovations/wp/2016/10/03/google-translate-is-getting-really-really-accurate/}, accessed February~20, 2018.} Deep learning has not only shown great success in natural language processing but also in image classification, object detection, object localization and image generation. For instance, Alipay introduced a mobile payment app to more than 120~million people in China that allows them to use face recognition for payments. This technology was ranked by \emph{Technology Review} as one of the ten breakthrough technologies of 2017.\footnote{\emph{Technology Review} (2017): \textquote{10 breakthrough technologies: Paying with your face}. \url{https://www.technologyreview.com/s/603494/10-breakthrough-technologies-2017-paying-with-your-face/}, last accessed February~20, 2018.} Aside from these applications, deep learning has also been successfully applied to recommendation systems. In this regard, both Amazon and Netflix utilize deep neural networks for personalized product recommendations.

\begin{figure}[H]
\centering
\includegraphics[width=.5\linewidth]{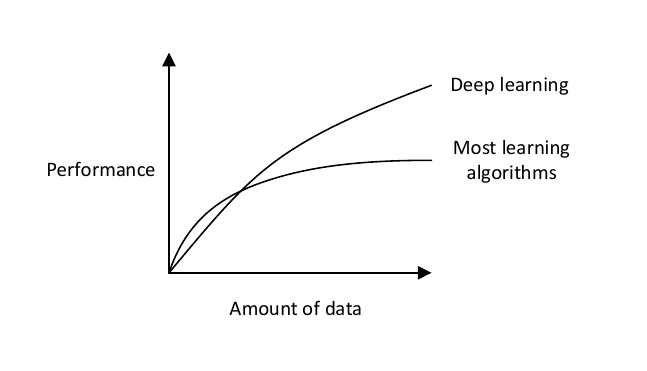}
\vspace{-1cm}
\caption{Illustrative comparison between performance of deep learning against that of most other machine learning algorithms. Deep neural networks still benefit from large amounts of data, whereas the performance increase of other machine learning models plateaus \citep{Ng.2016}}  
\label{fig:dataset_size}
\end{figure}

Why deep neural networks have only now become so powerful has several explanations \citep{Goodfellow.2017}:
\begin{enumerate}
\item \emph{Computational power.} Computational capabilities have increased rapidly, especially due to the widespread use of graphics processing units~(GPUs). These are particularly suited to executing the operations from linear algebra necessary for fitting neural networks. For instance, Google DeepMind optimized a deep neural network using \num{176} GPUs for \num{40} days to beat the best human players in the game Go \citep{Silver.2017}. This would have required far greater computational resources without acceleration through GPUs.
\item \emph{Data.} Second, large datasets are needed to train deep neural networks in order to prevent overfitting and fine-tune parameters (see \Cref{fig:dataset_size}). The performance of deep neural networks generally improves with increasing amounts of data; smaller datasets incorporating only several hundred of datapoints were not sufficient to optimize DNNs in previous years. However, in the era of big data, large datasets are now common in most businesses which are important, as empirical results show that even DNNs still benefit from additional data even when already having millions of datapoints \citep{Goodfellow.2017}. As a result, big data can be used effectively, often by mining public content from the Web. One prominent example, the so-called ImageNet dataset, was developed to support tasks in computer vision and comprises more than 14~million images. We later detail this aspect by showing how the size of the dataset affects the overall performance of deep learning. 
\item \emph{Optimization algorithms.} Optimizing the parameters in deep neural networks is a challenging undertaking. Several optimization algorithms have been proposed since \citet{Hinton.2006} published what is now regarded as the seminal work of deep learning. In this publication, the authors increase the depth of neural networks gradually by alternating between adding a new layer and optimizing the network parameters. This technique, which stabilizes the optimization, paved the way for learning deeper networks. Further innovations deal with the optimizer itself. For instance, the large size of datasets prohibits one from optimizing the overall performance directly; deep neural networks are instead trained by stochastic optimization \citep{LeCun.1998}. Here parameters are updated based on an approximation of the objective function (\ie by evaluating it on a subsample of the whole dataset). This distinguishes deep neural networks from many other machine learning models, which are not designed for big data and, hence, do not scale to millions of datapoints. To improve optimization, there are a number of common optimization methods (\eg Adam, Adagrad, RMSprop) that implement variants of stochastic gradient descent, often paired with an adaptive regulation of the step size. In addition, deep neural networks with millions of parameters suffered from overfitting the model to the training data. As a remedy, it is now recommended that one integrate regularization into the optimization procedure (\ie weight decay, dropout or batch normalization) in order to improve the generalizability.
\label{intro:explanations} 
\end{enumerate}

While deep learning is on the way to becoming the industry standard for predictive analytics within business analytics and operations research, our discipline is still in its infancy with regard to adopting this technology. To support this claim, we conducted an extensive literature review of all papers published by October 2018 across the premier journals in our field, namely, \emph{European Journal of Operational Research}, \emph{Operations Research}, \emph{Production and Operations Management}, \emph{Journal of Operations Management}, and \emph{Decision Sciences}. We specifically searched for papers containing the terms \textquote{deep neural network} or \textquote{deep learning}. Our search returned \num{15} matches but a closer inspection led us discard \num{12} of them, since these only contained the term in passing but without actually using deep learning. The resulting three matches are listed in \Cref{tbl:dl_literature}. This is an interesting observation as the operations research community has a longstanding tradition of applying neural networks. Examples of \mbox{1-hidden} layer networks appear in the areas of, for instance, healthcare \citep{Misiunas.2016, Oztekin.2017}, demand forecasts \citep{Carbonneau.2008, Venkatesh.2014} and maintenance \citep{Mahamad.2010,Mazhar.2007}. We further extended our literature search to other fields of management science, namely, accounting, finance, marketing and information systems; yet these outlets (as per Financial Times 50 list) published a total of only two papers that benefit from deep learning; see \citet{Adamopoulos.2018} and \citet{Li.2017}.

It is not only academia that has yet to fully incorporate deep learning in its decision-making routines. This fact also holds true for a majority of enterprises: a 2016 report on \textquote{The Age of Analytics} by the McKinsey Global Institute refers to deep learning as \emph{\textquote{the coming wave}} \citep{Henke.2016}. This is also borne out by our own expertise when collaborating with large-cap companies, including consulting firms, across Europe and the United States. One member from a top management consulting firm even admitted to us that \emph{\textquote{we don't have a clue how deep learning works, neither do our clients}}. Hence, the above mentioned showcases of deep learning are largely exceptions among a handful of selected firms, thereby highlighting the dire need for company professionals to better understand deep learning, its applications and value \citep[cf.][]{Lee.2018}. 

The objective of this overview is to provide an overview on the use of deep learning in academia and practice from an operations research perspective: (1)~we summarize the most relevant mathematical concepts in deep learning for readers who are not familiar with this technique. (2)~We demonstrate the use of deep learning across three case studies from operations research. Here we specifically compare its performance to conventional models from machine learning, thereby showing the improvements in both prediction and operational performance. As part of it, we see that default, out-of-the-box architectures are often not sufficient and, following this, we propose a novel deep-embedded network architecture. (3)~We provide recommendations regarding which network architecture to choose and how to tune parameters. (4)~We derive implications for managers and practitioners who want to engage in the use of deep neural networks for their operations or business analytics. (5)~We propose directions for future research, especially with regard to the use of deep learning for business analytics and operations research.

\begin{table}
\centering
\footnotesize
\makebox[\textwidth]{
    \begin{tabular}{lp{6cm}p{4cm}}
        \toprule
    	{\textbf{Paper}} & 
		{\textbf{Problem}} &
		{\textbf{\mcellt{Network\\ architecture}}} \\ 
        \midrule
		\citet{Krauss.2017} & Short-term arbitrage trading for the S\&P 500 & 5-layer perceptron \\[0.5cm]
		\citet{Probst.2017} & Generative neural networks for estimation of distribution algorithms & Stochastic neural network (\ie restricted Boltzmann machine)  \\
		\citet{Fischer.2018} & Predict directional movement of constituent stocks of the S\&P 500 & Long short-term memory \\
        \bottomrule
    \end{tabular}
}
\caption{Literature review for papers from operations research utilizing techniques from deep learning.}
\label{tbl:dl_literature}
\end{table}


The remainder of this overview article is structured as follows. \Cref{sec:mathematical_background} recapitulates deep neural networks by rewriting the concept as an optimization problem using conventional terminology of operations research, while \Cref{sec:methods} proposes our tailored deep-embedded network architecture. To demonstrate the potential use of deep learning in practice, we then conduct three experiments with real-world data from actual business problems (see \Cref{sec:computational_experiments}). All of our experiments reveal superior prediction performance on the part of deep neural networks. Finally, \Cref{sec:discussion} provides recommendations for the use of deep learning in operations research and highlights the implications of our work for both research and management. \Cref{sec:conclusion} concludes with a summary. 

\section{Mathematical background: From neural networks to deep learning}
\label{sec:mathematical_background}

This section reviews the transformation from \textquote{shallow} neural networks to deep learning. For a detailed description, we refer to \citet{Russell.2010} and specifically to \citet{Goodfellow.2017}. In addition, \citet{Schmidhuber.2015} presents a chronological review. 

\subsection{Predictive analytics}

Predictive models in business analytics exploit input features $\bm{x} \in \mathbb{X} \subseteq \mathbb{R}^m$ to predict an unknown variable $y \in \mathbb{Y}$. In practical settings, the input features could be recent sales figures in order to forecast future production needs, or historic sensor data to anticipate machinery failure. Depending on whether this is a discrete label ($y \in \{0, \ldots, k\}$) or a real value ($y \in \mathbb{R}$), we refer to it as a classification or regression task, respectively. The objective is then to find a mapping $f : \bm{x} \mapsto y$. 

The choice of such functions is given by the predictive model $f(\cdot; w)$ with additional parameters $w$; that is, $y \approx f(\bm{x}; w)$. Then, in practice, the objective behind the prediction results in an optimization problem whereby one must find the best parameters $w$. A variety of models $f$ are common in business analytics: examples from traditional machine learning involve, for instance, linear models \citep[e.\,g.][]{Bertsimas.2016,Bertsimas.2007}, decision trees, support vector machines, neural networks \citep[e.\,g.][]{Delen.2012,Oztekin.2016}, or even deep neural networks as motivated by this work. Each of them is often accompanied by a tailored optimization strategy; see \citet{Bertsimas.2014}.


The above optimization requires a performance measure that assesses the model, \ie the error between the predicted value $\hat{y} = f(\bm{x}; w)$ and the true observation $y$. This is formalized by a loss function $\mathcal{L} : \mathbb{Y} \times \mathbb{Y} \rightarrow \mathbb{R}$. The actual choice depends on the desired objective (\eg whether one wants to penalize false-positives or true-negatives) and the prediction task, \ie classification or regression. Predictive modeling now simplifies to minimizing the loss $\mathcal{L}(\hat y, y) = \mathcal{L}(f(\bm{x}; w), y)$, usually summed over a set of samples $i$. Hence, we yield the optimization problem
\begin{equation}
w^* = \argmin \limits_{w \in \mathbb{W}} \sum_{i} \mathcal{L}(\hat{y}_i, y_i) = \argmin \limits_{w \in \mathbb{W}} \sum_{i} \mathcal{L}(f(\bm{x}_i; w), y_i) ,
\label{equ:opt}
\end{equation}
where $\mathbb{W}$ denotes the weight space. With the increasing dimensionality of the weight space, the predictive model gains flexibility and thus becomes able to adapt to complex relationships between features and outcome. At the same time, a high-dimensional space heightens the computational requirements for solving the optimization problem. In addition, a large number of pairs $(\bm{x}, y)$ are required to ensure that there are sufficient samples with each combination of value, due to the curse of dimensionality \citep{Bellman.1972}.

The following sections address three issues: how to define the function $f$ in deep neural networks (as compared to traditional neural networks); how to find the parameters $w$; how to prevent overfitting.

\subsection{Neural networks}

\subsubsection{Single-layer neural networks}

\begin{figure}[H]
\centering
\includegraphics[width=.25\linewidth]{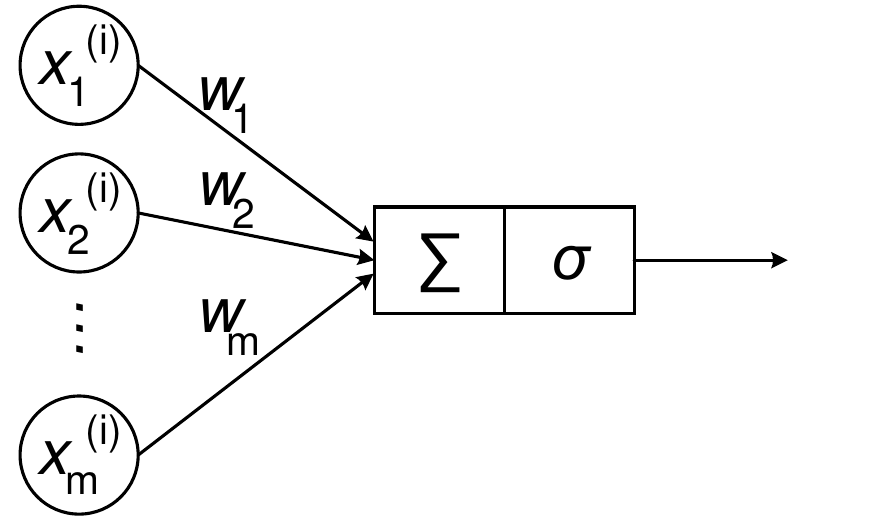}
\caption{Illustrative neuron in an artificial network with inputs $x^{(i)}_1,\dots,x^{(i)}_m$, weights $w_1,\dots,w_m$ and activation function $\sigma$. $\Sigma$ denotes the summation over all inputs.}
\label{fig:neuron}
\end{figure}

Similar to a biological neural network, an artificial neural network is a collection of connected units called neurons. An artificial neuron receives inputs from other neurons, computes the weighted sum of the inputs (based on weights $w$), and maps the sum via an activation function to the output. \Cref{fig:neuron} illustrates how a neuron receives inputs $x^{(i)}_1,\dots,x^{(i)}_m$ and processes them with weights $w_1,\dots,w_m$.

Finally, the output is passed to connected neurons as their inputs. If all connections follow a forward structure (\ie from input neurons that process features of $\bm{x}$ to the output neurons), we call this a feedforward neural network. Accordingly, the network is free of cycles or feedback connections that pass information backwards. The latter are called recurrent neural networks and are discussed in \Cref{sec:recurrent_neural_networks}.


The simplest neural network follows the above concept and is thus represented by a single-layer perceptron where the network $f_\text{1NN}$ is computed via a linear combination embedded in an activation function $\sigma$. We yield  
\begin{equation}
f_\text{1NN}(\bm{x}; W, b) = \sigma(W \, \bm{x} + b)
\end{equation}
with parameters $W$ (the weight matrix) and $b$ (an intercept called bias). Here the flexibility of the network stems from choosing an activation that is non-linear. 


Common choices of activation functions are as follows: 
\begin{itemize}
\item the sigmoid function $\sigma(x) = \cfrac{1}{1 + \e{-x}} \in \left( 0,1 \right) $ , 
\item the hyperbolic tangent $\sigma(x) = \cfrac{\e{2x} - 1}{\e{2x} + 1} \in \left[ -1,1 \right]$, and 
\item the rectified linear unit (ReLU) given by $\sigma(x) = \max(0,x) \in \left[ 0, \infty \right)$.
\end{itemize}
It is apparent that they all share a particular characteristic, namely, that a certain threshold must be exceeded in order for the activation functions to pass through values. Hence, the idea of activation functions is biologically inspired in the sense that they resemble neurons in the human brain, which also have to receive a certain stimuli in order to be activated \citep{Stachenfeld.2017}. In practice, the different choices entail their own (dis-)advantages; for instance, the ReLU activation function can lead to neurons which output zero for essentially all inputs, thereby losing flexibility. 


Because of the activation functions, neural networks can model non-linear and non-convex functions. However, this also makes the optimization problem denoted by \Cref{equ:opt} difficult to optimize. In particular, there exists no direct closed-form solution that gives the global optimum of the optimization problem. Instead, one uses gradient descent to find local optima. This technique has been shown to provide solutions that are close to the global optimum \citep{Choromanska.2015}. Mathematically, one updates parameter $w \in \mathbb{W}$ by calculating the partial derivative of $w$ with respect to the loss $\mathcal{L}$ of the training samples, \ie 
\begin{equation}
\sum_i \frac{\delta}{\delta w} \mathcal{L}(f(\bm{x}_i, w), y_i).
\end{equation}
Accordingly, one updates the parameter $w$ such that the loss $\mathcal{L}$ decreases via 
\begin{equation}
w \leftarrow w - \eta \sum_i \frac{\delta}{\delta w} \mathcal{L}(f(\bm{x}_i, w), y_i),
\label{equ:weight_update}
\end{equation}
where $\eta$ denotes the step size or learning rate of the gradient descent optimization. One proceeds similarly to update the bias $b$.

Single-layer neural networks have many limitations. Most famously, for monotonic activation functions, they cannot learn the XOR function given by $f_\text{1NN}([0,0],w) = 0$, $f_\text{1NN}([0,1],w) = 1$, $f_\text{1NN}([1,0],w) = 1$ and $f_\text{1NN}([1,1],w) = 0$. As a remedy, multi-layer neural networks consist of many layers that first transform their input into higher-dimensional representations and then into the output. In theory, the universal approximator theorem guarantees that neural networks with three layers are sufficient to represent arbitrary functions $f: \mathbb{X} \rightarrow{Y}$ \citep{Cybenko.1989}. Nevertheless, practical experience suggests that deeper models can better reduce the generalization error \citep{Goodfellow.2017}. 

\subsubsection{Deep neural networks}

While the previous neural networks consisted of only a single layer, one can extend the mathematical specification to multi-layered perceptrons. We refer to these using the term \textquote{deep}, which can reflect an arbitrary number of layers.\footnote{Although some architectures currently being used consist of only two to five hidden layers, only recent innovations in the realm of deep learning render it possible to make use of them. Hence, they are also considered as being deep neural networks.} In such networks, the number of free parameters increases, as well as the flexibility of the network to represent highly non-linear functions. We can formalize this mathematically by stacking several single-layer networks into a deep neural network with $k$ layers, \ie
\begin{equation}
f_\text{DNN}(\bm{x}) = \underbrace{f_\text{1NN}(f_\text{1NN}( \ldots f_\text{1NN}(\bm{x})))}_{k}
 = \underbrace{f_\text{1NN} \circ \ldots \circ f_\text{1NN}}_{k}(\bm{x}) .
\end{equation}
The first layer is referred to as the input layer, the last as the output layer and the remainder are termed hidden layers. 


We note that the dimension of each layer is not necessarily equal across all layers, that is to say it can differ. In practice, the depth of deep neural networks varies across applications, ranging between two hidden layers to potentially several hundred. Together with the corresponding dimension of each layer, this can easily result in networks that entail tens of millions of degrees of freedom. As a result, optimizing the weights in a deep neural network is a daunting task, requiring (1)~gradient-based optimization methods and (2)~regularization. Both are detailed in \Cref{sec:estimation}. 


The task of choosing an adequate number of layers and the number of neurons in each layer represents a challenging undertaking. \Citet{Montavon.2012} recommend adding layers until the generalization error stops improving. Other best practices suggest adding layers until the predictive model overfits on the training data and then removing the overfitting by regularization methods. Moreover, a large dimension of neurons in each layer is generally preferred, as these seldom interfere with the generalization error. Yet, we note that optimizing these hyperparameters is still subject to active research.


For deep neural networks, the activation function is commonly set to the rectified linear unit \citep{LeCun.2015}. This choice leads to sparse settings whereby a large portion of hidden units are not activated, thus having zero output. On the other hand, the recurrent network architectures (\cf \Cref{sec:recurrent_neural_networks}) are frequently utilized with sigmoid activation functions, since these constrain the output to between $0$ and $1$. Thereby, so-called gates can be defined which circumvent exploding gradients during the numerical optimization.


In classification tasks, the traditional loss functions, such as $L_1$ and $L_2$, suffer from slow learning. That is, the partial derivatives of the loss function are small as compared to a large value of the loss itself. Accordingly, classification tasks generally output a discrete probability distribution over all possible classes by using a softmax activation in the output layer \citep{Russell.2010}. However, for large losses, the partial derivative for traditional losses vanishes and, as a consequence, one typically prefers the cross entropy to measure the similarity between the output distribution and the target distribution \citep{Goodfellow.2017}. In regression tasks, neural networks output a continuous values, for which one frequently draws upon the mean squared error or the smoothed mean absolute error.  

\subsection{Model estimation}
\label{sec:estimation}

\subsubsection{Weight optimization}

The optimization in deep learning is analogous to the general setting in predictive analytics, the loss is minimized via
\begin{align}
\label{eqn:optimization_DNN}
w^* = \argmin\limits_{w\in\mathbb{W}} \mathcal{L} \left( f_\text{DNN}(\bm{x}; w), y \right) .
\end{align}
While the weights in a simple perceptron can be identified through convex optimization, the optimization problem for deep neural networks is computationally challenging due to the high number of free parameters. In fact, \citet{Judd.1990} proves that it is NP-hard to optimize a neural network so that it produces the correct output for all the training samples. That study also shows that this problem remains NP-hard even if the neural network is only required to produce the correct output for two-thirds of the training examples. 


The conventional solution strategy for optimizing deep neural networks involves gradient-based numerical optimization \citep{Saad.1998}. Similar to optimizing single-layer neural networks, one computes the partial derivatives of the parameters with respect to the loss $\mathcal{L}$ and changes the parameters in order to decreases the loss (\cf \Cref{equ:weight_update}). However, this must be done for all layers, from the output back to the input layer. Hence, a technique called backpropagation is preferable for reasons of efficiency \citep{Rumelhart.1986}. Backpropagation exploits the chain rule to reuse computations it has already calculated for the previous layer. Since these operations involve simple matrix operations from linear algebra, they can be run in parallel, turning GPUs into an effective accelerator for optimizations in deep learning. 


The runtime for optimization of a deep neural network can still be very high, since a single update of the parameters requires predictions $\hat y_i$ of all samples $\bm{x}_i$. As a remedy, stochastic gradient descent approximates the loss function across all samples with the loss of a smaller subset, called the \textquote{minibatch}. The size of the minibatch presents another hyperparameter, for which we typically recommend values between \num{32} and \num{256} based on our experience.  


The learning rate $\nu$ in \Cref{equ:weight_update} controls the step size during the optimization process. When this learning rate decreases at an appropriate rate, then stochastic gradient descent is guaranteed to converge to the global optimum under mild mathematical assumptions \citep{Kiwiel.2001}. Hence, the learning rate poses another hyperparameter that is usually chosen by visual inspection of the the learning curve, which traces the loss $\mathcal{L}$ as a function of time. Too high a learning curve coincides with oscillations, whereas a learning rate that is too low results in a slow optimization. To facilitate the choice, one may utilize an early stopping technique, which terminates the optimization when no improvement is achieved on the validation set for a certain time period.


Even though stochastic gradient descent is popular for optimizing deep neural networks, its performance for training can be very slow when the direction of the gradients changes (similar to a second-order derivative). A technique called momentum helps solve this issue by adding a velocity vector to the gradient \citep{Goodfellow.2017}. It accumulates a moving average of past gradients. As a result, momentum follows the drift of past gradients. 


The above mentioned concepts have been integrated into tailored optimization routines for deep learning. This has resulted in a variety of optimizers, such as as Adam (often considered a baseline), Adagrad, Adadelta and RMSProp, which are common in practice \citep{Goodfellow.2017}. The question of which algorithm performs best is still to be further studied. 

\subsubsection{Regularization}
\label{sec:regularization}

Optimizing deep neural networks typically requires a trade-off: on the one hand, one aims at a high number of free parameters as this allows for the representation of highly non-linear relationships. On the other hand, this makes the network prone to overfitting. The following remedies (sometimes used in combination) are common in deep learning and present forms of regularization to the weights:
\begin{enumerate}
\item \emph{Weight decay} adds a regularization term to the loss function, penalizing large weights in the network. With $W$ denoting the set of all parameters, the loss function thus changes to
\begin{equation}
\mathcal{L}_\text{WD} = \sum_i \frac{\delta}{\delta w} \mathcal{L}(f(\bm{x}_i, w), y_i) + \frac{\lambda}{2} \norm{W}_2.
\end{equation} 
Consequently, the gradient descent utilizes a new update rule given by
\begin{equation}
w \leftarrow w - \eta \left( \sum_i \frac{\delta}{\delta w} \mathcal{L}(f(\bm{x}_i, w), y_i) + \lambda \norm{W}_2 \right).
\end{equation}
As a result, the decision boundaries become smoother, thereby facilitating generalization of the network \citep{Goodfellow.2017}. 
\item \emph{Dropout} discards a small but random portion of the neurons during each iteration of training \citep{Srivastava.2014}. The underlying intuition is that the several neurons are likely to model the same non-linear relationship simultaneously; dropout thereby prevents neurons from co-adapting to the same features. Mathematically, this can be achieved by setting the corresponding rows in the weight matrix $W$ to zero.  
\item \emph{Batch normalization} (not to be mistaken with the previous updating in minibatches) performs a normalization of the output of each layer before forwarding it as input to the next layer \citep{Ioffe.2015}. As a result, this shrinks values closer to zero and, in practice, allows one to use higher learning rates with less care concerning the initialization of parameters. 
\end{enumerate}

\subsection{Advanced architectures}

Beyond the multi-layered perceptron discussed above, an array of alternative architectures have been proposed, often targeting specific data structures. \Citet{Goodfellow.2017}, as well as \citet{Schmidhuber.2015}, offer comprehensive overviews, while we summarize the most widely utilized choices in the following: convolutional networks that are common in vision tasks and recurrent networks for handling sequential data (see \Cref{tbl:overview_networks}). These can also be combined with the previous dense layers as additional building blocks. 

\begin{table}[H]
\centering
\footnotesize
\makebox[\textwidth]{
\begin{tabular}{p{3cm}p{3cm}p{4cm}p{4cm}}
\toprule
\textbf{\mcellt{Architecture}} & \textbf{Data structure} & \textbf{\mcellt{Examples}} & \textbf{\mcellt{Tuning parameters}} \\
\midrule
\multicolumn{4}{c}{\textsc{Dense neural networks}} \\
\midrule
Multi-layer perceptron (MLP) & Feature vectors of fixed length & Can be simple replacement of traditional models & Activation function, number of layers and neurons \\
\midrule
\multicolumn{4}{c}{\textsc{Convolutional neural networks}} \\
\midrule
Convolutional neural network (CNN) & High-dimensional data with local dependencies & Image recognition, speech recognition & Number and width of convolution filters\\
\midrule
\multicolumn{4}{c}{\textsc{Recurrent neural networks}} \\
\midrule
Long short-term memory (LSTM) & Sequential data & Text classification, time series forecasting & Number of layers and neurons \\[0.2cm]
Gated recurrent unit (GRU) & Sequential data & Text mining, time series forecasting & Number of layers and neurons \\
\bottomrule
\end{tabular}
}
\caption{Overview of common network architectures in deep learning.} 
\label{tbl:overview_networks}
\end{table}

\subsubsection{Convolutional neural network}

\begin{figure}[H]
\centering
\includegraphics[width=.45\linewidth]{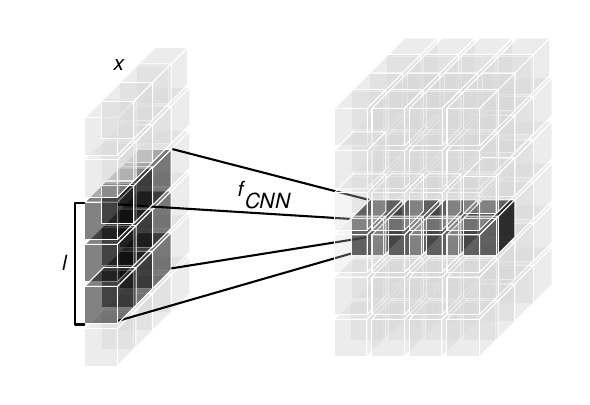}
\caption{Example of a convolution layer that exploits the spatial structure in the input data (left). It then applies a convolution operation to a local neighborhood in order to compute the output (right). Here $l$ denotes the kernel width.}  
\label{fig:CNN_illustration}
\end{figure}

The convolutional neural network~(CNN) exploits spatial dependencies in the data, \eg among neighboring pixels in an image. In other words, the idea of a CNN is to take advantage of a pyramid structure to first identify concepts at the lowest level before passing these concepts to the next layer, which, in turn, create concepts of higher level. Therefore, neurons are no longer connected with every other as a CNN goes from concepts that are identified locally to globally. Rather, neurons are now densely connected only in a small neighborhood (see \Cref{fig:CNN_illustration}). This choice is motivated by the human visual cortex, which also experiences stimulation in a restricted region of the visual field. 

Mathematically, the CNN applies a convolution operation (also known as kernels or filters) to the input and then passes the result to the next layer. Let  $\bm{x}_{ij}$ denote the element in row $i$ and column $j$ of a matrix $\bm{x} \in \mathbb{R}^{m \times m}$ that represents the values from the input layer. Then, the CNN calculates 
\begin{equation}
f_\text{CNN}(\bm{x})_{ij} = \sum_{x = -\frac{l}{2}}^{\frac{l}{2}} \sum_{y = -\frac{l}{2}}^{\frac{l}{2}} \bm{x}_{i + x, j + y} \, k_{xy},
\end{equation}
when convolving $\bm{x}$ with a kernel $k$ of width $l$. Here one usually pads $\bm{x}$ with zeros in order to avoid accessing non-existing values beyond the bounds of $\bm{x}$. The above operation can be analogously extended to higher-dimensional tensors. 

\subsubsection{Recurrent neural networks}
\label{sec:recurrent_neural_networks}


Traditional machine learning is limited to input vectors $\bm{x} \in \mathbb{R}^m$ of a fixed dimension $m$. This is rarely suited for sequences, which do not fit into such a structure of fixed size, since they entail varying lengths, ranging between 1 and an arbitrary number of elements. For this reason, sequence learning requires a problem specification whereby $f_\text{RNN} : \mathbb{X} \rightarrow \mathbb{Y}$ can handle input from $\mathbb{X} = \left\{ \mathbb{R}, \mathbb{R} \times \mathbb{R}, \mathbb{R}^3, \ldots \right\}$. In other words, this formalization takes sequences $\bm{x}_i = \left[ x^{(i)}_1, \ldots, x^{(i)}_{\tau_i} \right]$ as input, but each with its own length $\tau_i$. Prominent examples of sequential data in practical application include time series or natural language (where the running text is represented as a series of words or characters). 

\begin{figure}[H]
\centering
\includegraphics[width=.5\textwidth]{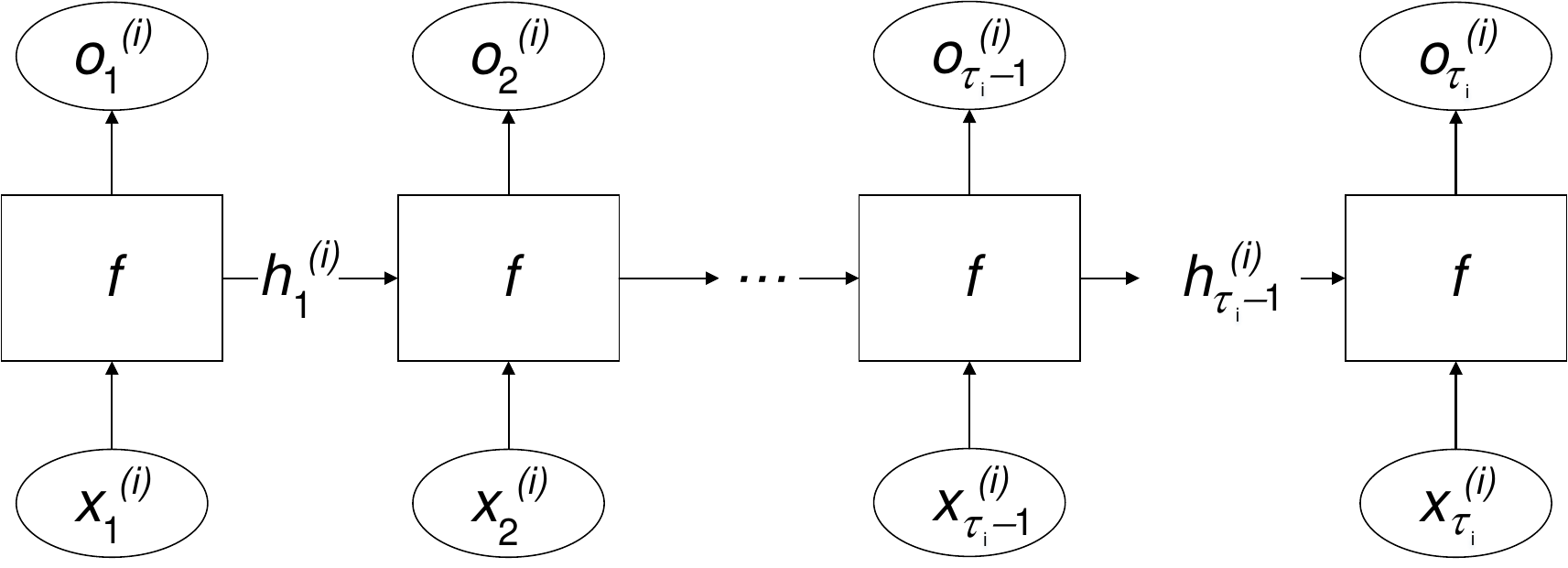}
\caption{Schematic illustration of a recurrent neural network. Here the same network $f$ is applied to each element $x^{(i)}_1, \ldots, x^{(i)}_{\tau_i}$ of the sequence. The network utilizes not only the current element $x^{(i)}_{k}$ of the sequence as input, but also the internals (\ie the hidden layer $h^{(i)}_{k-1}$) from the network belonging to the previous element in the sequence. Thereby, the knowledge of the whole sequence is accumulated in the hidden states. The hidden layers are denoted by $h^{(i)}_1, \ldots, h^{(i)}_{\tau_i-1}$ and the output of the neural network by $o^{(i)}_1, \ldots, o^{(i)}_{\tau_i}$.}
\label{fig:RNN_unrolled}
\end{figure}

Recurrent neural networks are specifically developed to handle such sequential input. They iterate over input sequences $\bm{x}_i$ without making assumptions regarding the length of the sequence. Hence, the same weights of the neural network are applied to process each element in the sequence; see \Cref{fig:RNN_unrolled}. Here the neural network not only processes the current element in the sequence, but also draws upon the hidden layer of the previous element in the sequence. As a result, the recurrent structure allows to pass information onwards while iterating over the sequence. This implicitly creates a \textquote{state} whereby information is stored and accumulated. It thus encodes the whole sequence in the hidden layer $h^{(i)}_\tau$ of the last neural network.

Mathematically, the network input for element $k \in \{ 1,\dots, \tau_i \}$ is given by a concatenation between the current element $x^{(i)}_k$ and the previous hidden state $h^{(i)}_{k-1}$, \ie $[x^{(i)}_k,h^{(i)}_{k-1}]$. The recurrent neural network thus computes
\begin{equation}
f_\text{RNN}(x_1, \ldots, x_\tau) = f([x^{(i)}_\tau, f([x^{(i)}_{\tau-1}, \ldots f([x^{(i)}_1] ; W, b);  \ldots ]; W, b  ]; W, b) ,
\end{equation} 
where the underlying network $f$ can be, for instance, a simple single-layer neural network or a deep one. In practice, long sequences result in a large number of recurrent calls. This causes the optimization to become numerically highly unstable \citep{Bengio.1994}. Hence, one might consider two extended network architectures that explicitly control how information is passed through:
\begin{itemize}
\item \emph{Gated recurrent unit (GRU)}. The gated recurrent unit utilizes two underlying neural networks, an update gate and a reset gate, in order to explicitly determine how values in the hidden states are computed.  Each of the gates is linked to an element-wise multiplication with values between 0 and 1, thus determining what ratio of each value is kept or discarded. 
\item \emph{Long short-term memory (LSTM).} In addition to the hidden state, this network structure builds upon an internal representation (called cell) which stores information \citep{Hochreiter.1997}. It further includes multiple feedforward layers that formalize how values in cell are erased (called forget gate), how the input is written to the cell (input gate), and how the cell computes output (output gate). 
\end{itemize}
The gated recurrent unit can be seen as a modification of the LSTM with fewer parameters. Yet, practical evidence suggests that both networks yield a comparable performance across a variety of tasks \citep{Chung.2014}.

\subsection{Embeddings}

Most machine learning models and especially deep neural networks struggle with input in the form of categorical variables \citep{Jia.2014}. The prime reasons are the mathematical properties of their one-hot encodings: one-hot encodings are (i)~non-continuous, (ii) ignore interdependencies and (iii)~their complexity scales with their cardinality. Different remedies have been suggested as follows. \citet{Zhang.2016} suggest the use of a factorization machine for pre-processing categorical variables (which serves as one of our benchmarks). \citet{Cerda.2018} develop similarity-based embeddings for handling noise in categorical variables (\eg variables where texts that have spelling errors). \citet{Jia.2014} present an embedding layer customized to categorical variables; however, this embedding layer is estimated via unsupervised pre-training: such estimation is not practical in most real-world OR applications where data is limited. In general, embeddings not only aid deep learning but also feature preprocessing in traditional machine learning \citep{Jia.2014,Zhu.2017}.

\section{Methods and materials}
\label{sec:methods}

\subsection{Preprocessing}
\label{sec:opt_preprocessing}


Deep neural networks entail an inherent advantage over traditional machine learning as they can handle data in its raw form without the need for manual feature engineering \citep{LeCun.2015}. In contrast, the conventional approach is to first devise rules and extract specific representations from the data. Examples are extracting edges from images instead of taking only the pixels as input, counting word triplets in natural language processing instead of processing the raw characters, or replacing individual time series observations with descriptive statistics such as minimum and maximum values. Deep learning largely circumvents the need for feature engineering and operates on the original data -- pixels, characters, words or whole time series.


In a variety of applications, the only compulsory preprocessing step is to replace categorical values with a numeric representation. One usually approaches this by utilizing a one-hot encoding: the categorical value with $K$ different entities is replaced by a $K$-dimensional vector where a $1$ in an element indicates that the corresponding category is active and will otherwise be $0$. Hence, the vector is $0$ almost everywhere except for a single $1$ that refers to the category.


One-hot vectors of large size lead to networks that become numerically difficult to optimize. An alternative is to replace sparse one-hot vectors with a so-called embedding, which maps  them onto low-dimensional but dense representations \citep{Goodfellow.2017}. Such embeddings are again modeled by a simple neural network and can even be optimized at the same time as the original deep neural network. Notably, embeddings can be constructed in such a way that they map from several neighboring input vectors onto the low-dimensional representation in order to encode additional semantics \citep{Hirschberg.2015}. When processing natural language, one can utilize pre-computed word embeddings such as word2vec or GloVe; however, these are often not available for domain-specific applications in OR.


\subsection{Proposed deep-embedded network architecture}

Given the challenges surrounding categorical variables, we suggest a neural network architecture that specifically addresses the need of effectively handling categorical variables in OR practice. For this reason, our architecture presents a combination of the abovementioned embedding layers and, in addition, stacked neural layers (such as MLPs, CNNs or RNNs). Our architecture is particularly helpful in dealing with data of inhomogeneous types, \ie consisting of both categorical and numerical input. Intuitively, our deep-embedded network architecture translates categorical variables into a dense representation, thus allowing for a numerically stable optimization of the neural network even for a large number of different categories. Mathematically, the architecture consists of two components: (i)~the embedding of categorical variables and (ii)~the concatenation of the embedded categorical variables with numerical variables in order to forward it to stacked neural layers. For both components, the optimization of all layers is performed simultaneously.
	
i) Let $N_\text{cat}$ denote the number of categorical variables $\mathcal{I}^{(i)}_\text{cat}, i=1,\dots,N_\text{cat}$ and let $\mathcal{I}_\text{num}$ denote numerical variables. An embedding layer $\mathcal{E}^{(i)}$ is utilized to map one categorical variable $\mathcal{I}^{(i)}_\text{cat}$ to a dense vector $v_i$ of dimension $d$ via an embedding layer, \ie
\begin{align}
v_i = \mathcal{E}_i(\mathcal{I}^{(i)}_\text{cat})
\end{align}
	
ii) Subsequently, all dense vectors $v_i, i=1,\dots,N_\text{cat}$ are concatenated along with $\mathcal{I}_\text{num}$ to one large vector $\mathcal{X}$, \ie
\begin{align}
\mathcal{X} = [v_1,\dots,{v_N}_\text{cat},\mathcal{I}_\text{num}].
\end{align}
Then, $\mathcal{X}$ yields input for default MLP, RNN, or CNN layers representing all input variables in dense form.

In order to counteract overfitting, the architecture is further extended by components representing dropout layers and batch normalization. In detail, batch normalization is added in all layers of the architecture up to the output layer, which utilizes batch normalization and dropout. Thereby, the non-linearity of the output layer is first applied before values are subject to batch normalization and dropout layers.

Our architecture is highly flexible: it can be utilized along MLPs, CNNs and RNNs that yield the second component of hidden and output layers. This is later shown based on different case studies involving a deep-embedded DNN and a deep-embedded LSTM. \Cref{alg:deepembedded} provides pseudocode of the deep-embedded network architecture utilized for our case studies.

\begin{algorithm}
\begin{minipage}[t]{0.46\textwidth}	
	\footnotesize
	\begin{algorithmic}[1]
		\Procedure{DeepEmbeddedDNN}{}
		\State \textbf{Inputs: }
		\State \quad - Vector of categorical features $\mathcal{I}_\text{cat}$
		\State \quad - Vector of numerical features $\mathcal{I}_\text{num}$
		\State \quad - Number of categorical features $N_\text{cat}$
		\State \quad - Number of numerical features $N_\text{num}$
		\State \quad - Dimension of embeddings $d$
		\State \quad - Number of hidden units $h$
		\State \quad - Number of layers $l$
		\State \quad - Dropout probability $p$ \\
		\For{$i$ \textbf{in} $1,\ldots,N_\text{cat}$}
		\State $n \gets \text{number of different categories}$
		\State \quad \quad \text{ of feature }  $\mathcal{I}_\text{cat}[i]$ 
		\State $v_i \gets \text{EmbeddingLayer}(n, d)(\mathcal{I}_\text{cat}[i])$ 
		\State $v_i \gets \text{Dropout}(v_i)$
		\EndFor 
		\State $\mathcal{X} \gets \text{Concatenate}(v_1,\dots,{v_N}_\text{cat}, \mathcal{I}_\text{num})$ 
		\State $\mathcal{X} \gets \text{FeedforwardLayer}(N_\text{num} + N_\text{cat} * d, h)(\mathcal{X})$ 
		\For{$i$ \textbf{in} $1,\ldots,l$}
		\State $\mathcal{X} \gets \text{FeedforwardLayer}(h, h, ReLU)(\mathcal{X})$ 
		\State $\mathcal{X} \gets \text{BatchNormalization}(\mathcal{X})$
		\EndFor 
		\State $\mathcal{X} \gets \text{Dropout}(\mathcal{X})$
		\State $\mathcal{X} \gets \text{FeedforwardLayer}(h, 1)(\mathcal{X})$ 
		\EndProcedure
	\end{algorithmic}
\end{minipage}
\hfill
\begin{minipage}[t]{0.46\textwidth}
	\footnotesize
	\begin{algorithmic}[1]
		\Procedure{DeepEmbeddedLSTM}{}
		\State \textbf{Inputs: }
		\State \quad - Matrix of categorical features $\mathcal{I}_\text{cat}$
		\State \quad - Number of categorical features $N_\text{cat}$
		\State \quad - Number of numerical features $N_\text{num}$
		\State \quad - Dimension of embeddings $d$
		\State \quad - Number of hidden units $h$
		\State \quad - Number of LSTM layers $k$
		\State \quad - Number of feedforward layers $l$
		\State \quad - Dropout probability $p$ \\
		\For{$i$ \textbf{in} $1,\ldots,N_\text{cat}$}
		\State $n \gets \text{number of different categories}$
		\State \quad \quad \text{ of feature }  $\mathcal{I}_\text{cat}[i]$ 
		\State $v_i \gets \text{EmbeddingLayer}(n, d)(\mathcal{I}_\text{cat}[i])$ 
		\State $v_i \gets \text{Dropout}(v_i)$
		\EndFor 
		\State $\mathcal{X} \gets \text{Concatenate}((v_1,\dots,{v_N}_\text{cat}, \mathcal{I}_\text{num})$ 
		\State $\mathcal{X} \gets \text{FeedforwardLayer}(N_\text{num} + N_\text{cat} * d, h)(\mathcal{X})$ 
		\For{$i$ \textbf{in} $1,\ldots,k$}
		\State $\mathcal{X} \gets \text{LSTM layer}(h)(\mathcal{X})$ 
		\EndFor
    	\State $\mathcal{X} \gets \text{Last hidden state of LSTM}$ 
		\For{$i$ \textbf{in} $1,\ldots,l$}
		\State $\mathcal{X} \gets \text{FeedforwardLayer}(h, h, \mathrm{ReLU})(\mathcal{X})$ 
		\State $\mathcal{X} \gets \text{BatchNormalization}(\mathcal{X})$
		\EndFor  
		\State $\mathcal{X} \gets \text{Dropout}(\mathcal{X}, p)$
		\State $\mathcal{X} \gets \text{FeedforwardLayer}(h, 1)(\mathcal{X})$  
		\EndProcedure
	\end{algorithmic}
\end{minipage}
\caption{}
\label{alg:deepembedded}
\end{algorithm}

Our work differs from the embeddings that were used in prior literature \citep{Jia.2014,Zhu.2017}: (1)~instead of using pre-training, our embeddings are directly integrated in the neural network architecture and are thus trained jointly. This aids a use in OR settings with scarce data. (2)~In our work, we also integrate embeddings into a LSTM architecture where both LSTM and embeddings are trained jointly. 
 (3)~Joint training introduces the risk of overfitting and, as a remedy, we refrain from using the na{\"i}ve embedding layer (this only serves as a baseline). Instead, we suggest the use of a dropout-based embedding layer. The na{\"i}ve embeddings later represent one of our baselines. 

\subsection{Estimation details}

Following general conventions, we tune all parameters in traditional machine learning using 10-fold cross-validation and reduce the computational time for deep learning by selecting \SI{10}{\percent} of the training samples for validation (\ie a random sample for case study~1 and chronological splits for the remaining time series predictions). We report our tuning ranges for each hyperparameter in the online appendix.

Given $n$ different categories of a categorical variable, we calculate the dimension $d$ of the embedding layer following common guidelines\footnote{\emph{Google Developers} (2017): \textquote{Introducing TensorFlow Feature Columns}. \url{https://developers.googleblog.com/2017/11/introducing-tensorflow-feature-columns.html}, last accessed February~20, 2018.} via $d = \sqrt[4]{n}$.

The required times for optimizing all model parameters are listed in the online appendix. As we shall see in the following, a higher optimization time for neural networks is countervailed by a higher prediction performance. We also provide loss curves for our deep-embedded networks in the online appendix.

\subsection{Experimental setup}

We perform a series of computational experiments in order to demonstrate the added benefit of deep learning in business analytics. For this purpose, we draw upon three case studies with public and proprietary industry data from different business areas as outlined in \Cref{tbl:datasets}. As part of our benchmarks, we utilize common methods from traditional machine learning, namely, linear models (Lasso and ridge regression), a tree-based approach (\ie the random forest) and further non-linear variants (\ie a support vector machine and a single-layer neural network). In addition, our experiments build upon common techniques for preprocessing categorical variables from prior literature, namely a factorization machine \citep{Zhang.2016} and deep neural networks with na{\"i}ve embedding layers \citep{Jia.2014}. Performance is compared primarily based on the mean squared error for regression and the area under the curve for classification tasks.\footnote{We refrain from using percentage errors as seen in some Kaggle competitions: there are crisp mathematical limitations \citep{Tofallis.2015} and these metrics do not measure costs. Instead, we evaluate all models through the lens of OR practitioners and thus report cost metrics. }

\begin{table}[H]
\centering
\footnotesize
\makebox[\textwidth]{
    \begin{tabular}{p{2.0cm}lp{2.0cm}p{1.2cm}p{3.7cm}p{4.0cm}}
        \toprule
    	{\textbf{Business area}} & 
    	{\textbf{Samples}} &
		{\textbf{\mcellt{Type of\\DNN}}} & 
		{\textbf{\mcellt{Public/\\Private}}} & 
		{\textbf{\mcellt{Categorical\\variables}}} & 
		{\textbf{Description}} \\
        \midrule
        Operations management & 273,750 & GRU/LSTM &  Private & Holiday flag & Predict the number of helpdesk tickets for the following hour\\[0.2cm] 
        Inventory management & 1,017,209 & GRU/LSTM & Public & Store ID, state holiday, school holiday, store type, assortment & Predict the sales volume on day-ahead of pharmacy stores \\[0.2cm] 
        Risk management & 595,212 & MLP & Public & 14 Anonymized variables & Predict whether a customer will file a claim for an automotive insurance company \\
        \bottomrule
    \end{tabular}
}
\caption{Overview of our case studies.}
\label{tbl:datasets}
\end{table}

\section{Numerical results}
\label{sec:computational_experiments}

\subsection{Case study 1: Load forecasting of IT service requests}
\label{sec:case_2}

The second case study focuses on operations management, where the number of incoming service requests for an IT department is predicted in order to better adapt available capacities to the load. An accurate prediction of workloads is essential in order to coordinate employees in IT services \citep{Bassamboo.2009}, as well as to avoid long waiting times for customers. For this purpose, we forecast the hourly number of tickets for the next work day based on historic values. 


This experiment draws upon a proprietary dataset of helpdesk tickets from a Swiss organization with approximately 10,000 employees. Our dataset spans January 2011 to mid-November 2017, totaling 267,397 tickets. On average, \num{5.163} tickets are sent per day with a peak value of \num{547}. In the first step, we counted the number of tickets in each hour of the study horizon. In total, this results in 60,192 samples. We augmented this dataset with an additional categorical variable that indicates holidays if there is currently a holiday. Our objective requires that we take a sequence of features $x_1,\ldots,x_{\tau}$ as input that represent the workload at each hour. We then predict the workload of the following day $y_{\tau+1}, \ldots,y_{\tau+24}$ across each hour.


\Cref{tbl:results_tickets} compares the prediction performance of traditional machine learning with that of deep learning. The average ticket volume per hour (\ie \num{3.7} tickets/h) is used as a na{\"i}ve baseline. Among the baseline models, we find the lowest mean absolute error on the test set when using a support vector machine, whereas a default deep neural network (two layers, each having 32 neurons) shows lowest mean squared error across the baseline models. The latter yields an improvement of \SI{42.95}{\percent} compared to mean value as the predictor. Our deep-embedded architecture outperforms all baseline models including the default neural network. It yields an improvement of \SI{3.75}{\percent} as compared to the best performing baseline, totaling in an improvement of \SI{45.10}{\percent} points over the mean value as the baseline predictor. Statistical significance tests on the mean absolute error demonstrate that deep neural networks outperform our baselines to a statistically significant degree at the \SI{1}{\percent} level. We also assessed the cost effectiveness in which prediction errors incur costs due to either idle workers or unsatisfied customers. Here our deep learning approach suggests savings of \num{29658.0} monetary units.

\begin{table}[H]
\centering
\scriptsize
\makebox[\textwidth]{
    \begin{tabular}{l S[table-format=7.0,group-separator={,}] SS SS}
        \toprule
    	{\textbf{Model}} & 
    	{\textbf{\mcellt{Free parameters}}} &
		\multicolumn{4}{c}{\textbf{Out-of-sample performance}} \\
		\cmidrule(l){3-6} 
		& & 
		{\textbf{MSE}} &
        {\textbf{\mcellt{MAE}}} &
		\textbf{\mcellt{Explained\\ variance ($R^2$)}} &
		\textbf{\mcellt{Prediction\\ error costs}} \\ 
        \midrule
		\multicolumn{2}{l}{\textsc{Baselines}} \\
		\quad Mean value as predictor & {---} & 191.157 & 7.931 & 0.000 & 46124.6 \\  
		\quad ARMA/ARIMA & 100 & 112.670 & 5.766 & 0.349 & 32454.4 \\
        \quad Lasso & 3649 & 115.983 & 5.842 & 0.324 & 32634.1 \\      
        \quad Ridge regression & 3649 & 114.800 & 5.940 & 0.331 & 32663.2 \\   
        \quad Random forest & \multicolumn{1}{c}{n/a} & 121.790 & 5.544 & 0.290 & 33015.0 \\ 
        \quad Support vector machine & 3649{$^\dagger$} & 129.110 & 5.355 & 0.250 & 32147.1 \\ 
		\quad Default neural network (single-layer) & 3649 & 112.197 & 5.650 & 0.346 & 33445.9 \\
		\quad Default DNN (tuned depth) & 374902 & 109.040 & 5.785 & 0.365 & 33136.1 \\
        \quad LSTM with factorization machine & 14596 & 121.782 & 5.458 & 0.322 & 32893.3 \\
        \quad LSTM with embeddings & 11672 & 121.262 & 5.567 & 0.286 & 33091.3 \\
        \midrule
        \multicolumn{2}{l}{\textsc{Proposed network architecture}} \\
        \quad Deep-embedded LSTM & 11672 & \bfseries 104.952 & \bfseries 5.101 & \bfseries 0.390 & \bfseries 29658.0 \\ 
        \bottomrule
\multicolumn{4}{l}{$^\dagger$ We utilize a kernel approximation along with a linear support vector machine to reduce computation time.} 
    \end{tabular}
}
\caption{Numerical results for hourly prediction of incoming service requests. Here the prediction performance is compared between common baselines from traditional machine learning and sequence learning with deep neural networks. Reported are the mean squared error (MSE), the mean absolute error (MAE) on the original unscaled values, the explained variance ($R^2$) and the cost from prediction errors (where both an overestimated and an underestimated volume incur costs of \num{1.0} per ticket due to workers being idle or customers being unsatisfied). The best performance is highlighted in bold.}
\label{tbl:results_tickets}
\end{table}

\subsection{Case study 2: Sales forecasting}
\label{sec:case_3}

Sales forecasting facilitates the decision-making and planning within firms \citep{Lau.2018, Boone.2018}. Hence, this case study evaluates deep learning for sales forecasting, where the one-day-ahead sales volume for each store is predicted based on the history of previous sales. The dataset comprises 842,806 days of past sales from 1113 different pharmacies.\footnote{For reasons of reproducibility and comparability, we chose a public dataset from \url{https://www.kaggle.com/c/rossmann-store-sales.}} In terms of preprocessing, we generate the following categorical variables that are embedded into a dense representation by our deep-embedded architecture: store ID, state holiday, school holiday, store type, assortment. In addition, we utilize the following numerical variables: distance to the next competitor, day of week and week of year.\footnote{These features were selected from the original dataset for the following reasons: some features are company-specific, whereas our selection should be widely applicable to other companies. Further, these features entail the largest predictive power with respect to the target variable and should thus be decisive.} The average sales volume numbers to \num{6954.97} which presents our na{\"i}ve baseline. 


\Cref{tbl:results_sales} lists the prediction performance of traditional machine learning, the default DNNs and our deep-embedded neural network. Among the baseline models, we find the lowest mean squared error on the test data when using a default DNN architecture (with three layers, each having 64 neurons). This model shows an improvement of \SI{86.49}{\percent} in terms of mean squared error. Our deep-embedded LSTM decreases the mean squared error by \SI{21.73}{\percent}. These improvements are also statistically significant at the \SI{1}{\percent} level as shown by a $t$-test. In total, our deep learning model yield an impressive improvement of up to \SI{89.42}{\percent} compared to predicting the mean value. The estimated costs from prediction errors again support the effectiveness of using deep learning. Altogether, similar to the first case study, our deep-embedded neural network outperforms both traditional machine learning models and default neural networks.

\begin{table}[H]
\centering
\scriptsize
\makebox[\textwidth]{
    \begin{tabular}{l S[table-format=9.0,group-separator={,}]  SS SS}
        \toprule
    	{\textbf{Model}} & 
    	{\textbf{\mcellt{Free parameters}}} &
		\multicolumn{4}{c}{\textbf{Out-of-sample performance}} \\
		\cmidrule(l){3-6} 
		& & 
		{\textbf{MSE}} &
        {\textbf{\mcellt{MAE}}} & 
		\textbf{\mcellt{Explained\\ variance ($R^2$)}} & 
		\textbf{\mcellt{Prediction\\ error costs}} \\
        \midrule
		
		\multicolumn{2}{l}{\textsc{Baselines}} \\
		\quad Mean value as predictor & {---} & 9979310.0 & 2283.6 & 0.000 & 190565826.7 \\ 
		\quad ARMA/ARIMA & 50 & 2225451.0 & 1024.1 & 0.773 & 83949310.9 \\
        \quad Lasso & 512 & 2321334.6 & 1045.6 & 0.764 & 84511640.9 \\     
        \quad Ridge regression & 512 & 2224958.3 & 1071.8 & 0.774 & 84212283.9 \\   
        \quad Random forest & \multicolumn{1}{c}{n/a} & 1376444.2 & 805.7 & 0.860 & 64960715.8 \\ 
        \quad Support vector machine & 512{$^\dagger$} & 2453320.3 & 1130.6 & 0.751 & 84908542.3 \\ 
		\quad Default neural network (single-layer) & 512 & 1871942.3 & 968.7 & 0.820 & 71928430.8 \\ 
        \quad Default DNN (tuned depth) & 61203 & 1348397.9 & 825.3 & 0.844 & 64891027.0 \\ 
        \quad LSTM with factorization machine & 3072 & 2432206.1 & 1185.1 & 0.753 & 95563305.5 \\
        \quad LSTM with embeddings & 10849 & 1609544.4 & 925.3 & 0.837 & 65983043.1 \\
        \midrule
        \multicolumn{2}{l}{\textsc{Proposed network architecture}} \\
        \quad Deep-embedded LSTM & 10849 & \bfseries 1055438.7 & \bfseries 713.6 & \bfseries 0.893 & \bfseries 57072820.5 \\
        \bottomrule
        \multicolumn{4}{l}{$^\dagger$ We utilize a kernel approximation along with a linear support vector machine to reduce computation time.} 
    \end{tabular}
}
\caption{Numerical results for predicting sales. The text dataset is kept the same as in previous experiments, yet we vary the size of the training set, \ie we take a subset of our original set in order to study the sensitivity of deep learning to large-scale datasets. Here the prediction performance is compared between common baselines from traditional machine learning and sequence learning with deep neural networks. Reported are the mean squared error (MSE), the mean absolute error (MAE)  on the original unscaled values, the explained variance ($R^2$) and the cost from prediction errors (where an overestimated volume incurs costs of \num{0.5} monetary units due to unsold products (incl. storage, etc.) and where an underestimated volume comes with penalty of \num{1.0} due to unsatisfied customers). The best performance is highlighted in bold.}
\label{tbl:results_sales}
\end{table}

\subsection{Case study 3: Insurance credit scoring}
\label{sec:case_1}

Our third case study relates to credit risk analysis \citep{Bassamboo.2008}, as it aims at identifying insurance credit scores. This translates into computing the likelihood of a customer filing a claim within a given time period. Based on these predictions, an insurance company can adapt their risk scoring and charge a corresponding premium. In general, predictive models found widespread application for tasks related to risk scoring \citep{Lessmann.2015}.


This dataset comprises 595,212 real-world customer records with \num{43} numerical and \num{14} categorical covariates provided by an insurance company for automotives.\footnote{For reasons of reproducibility and comparability, we experiment with a dataset from a public data science competitions (\url{https://www.kaggle.com/c/porto-seguro-safe-driver-prediction}). However, the participating models, as well as the test set of the competition, are confidential and we thus cannot compare our model to the deep neural network of winning contestant.} The variable that is to be predicted is whether a customer would file an insurance claim in the subsequent year. Hence, we obtain a classification task with two classes: filed and not filed.

Given the inherent nature of risk modeling, the dataset is highly unbalanced, as only \SI{3.6}{\percent} of the customers filed a claim. For this reason, we assess models in terms of the AUC of the receiver operating characteristic and the Gini coefficient, both of which account for imbalanced classes in the dataset.


\Cref{tbl:results_insurance_claims} compares the prediction performance. Among the baseline models from traditional machine learning, we find the highest Gini coefficient and the highest AUC score on the test data when using the ridge regression. This model shows an improvement of \SI{26}{\percent} to the majority vote. 

The deep-embedded DNN architecture (tuned to two hidden feedforward layers) outperforms all baseline models. The model increases the AUC score by \num{0.01} (\ie \SI{1.56}{\percent}) and the Gini coefficient by \num{0.02} (\ie \SI{7.14}{\percent}) as compared to the best-performing traditional machine learning model. Although the two-layer network is fairly shallow, deeper architectures did not result in a higher performance in this case. This can occur when the relationship between dependent variables and predicted outcome is fairly linear. Statistical tests on the AUC show that the improvement of the deep learning models is significant at the \SI{1}{\percent} level. Altogether, our deep-embedded model can successfully improve predictive accuracy over both traditional models and out-of-the-box DNN architectures, thereby allowing for a more precise identification of insurance scores and thus premiums. This finding is additionally established when assessing the cost effectiveness, where, given the assumed numbers, deep learning results in a misclassfication cost of only \num{19110.2} monetary units.

\begin{table}[H]
	\centering
	\scriptsize
	\makebox[\textwidth]{
		\begin{tabular}{l S[table-format=8.0,group-separator={,}]  SSS S}
			\toprule
			{\textbf{Model}} & 
			{\textbf{\mcellt{Free parameters}}} &
			\multicolumn{2}{c}{\textbf{Out-of-sample performance}} \\
			\cmidrule(l){3-4}
			& & 
			{\textbf{Gini}} &
			{\textbf{AUC}} & 
			\textbf{\mcellt{Misclassification\\ costs}}\\ 
			\midrule
			\multicolumn{2}{l}{\textsc{Baselines}} \\
			\quad Majority vote (\ie no claim) & {---} & 0.0 & 0.5 & 21830.0 \\   
			\quad Lasso & 272 & 0.254 & 0.627 & 19827.1 \\   
			\quad Ridge regression & 272 & 0.260 & 0.630 & 19343.2 \\ 
			\quad Random forest & \multicolumn{1}{c}{n/a} & 0.0 & 0.5 & 21830.0 \\ 
			\quad Support vector machine & 272{$^\dagger$} & 0.0 & 0.5 & 21830.0 \\     
			\quad Default neural network (single-layer) & 272 & 0.0 & 0.5 & 21830.0 \\   
			\quad Default DNN (tuned depth) & 37203 & 0.249 & 0.625 & 19931.1 \\
			\quad DNN with factorization machine & 1360 & 0.259 & 0.629 & 19958.1 \\
			\quad DNN with embeddings & 177745 & 0.232 & 0.616 & 20981.3 \\
			\midrule
			\multicolumn{2}{l}{\textsc{Proposed network architecture}} \\
			\quad Deep-embedded DNN  & 177745 & \bfseries 0.280 & \bfseries 0.640 & \bfseries 19112.0 \\
			\bottomrule
			\multicolumn{4}{l}{$^\dagger$ We utilize a kernel approximation along with a linear support vector machine to reduce computation time.} 
		\end{tabular}
	}
	\caption{Numerical results for determining insurance credit scores. Here the prediction performance is compared between common baselines from traditional machine learning and deep neural networks. Misclassification costs are computed based on an average loss of \num{0.1} monetary units per false positive and a lost profit of \num{10.0} for false negative.} The best performance is highlighted in bold.
	\label{tbl:results_insurance_claims}
\end{table}

\subsection{Sensitivity to size of training set}

The aforementioned case studies show that deep neural networks outperform traditional machine learning methods. This section shows results of further experiments that (1) compare the performance of deep learning and traditional machine learning for a variety of training observations and (2) give insights to the training process of deep neural networks. 

\begin{figure}[H]
\centering
\makebox[0.6 \textwidth]{%
\includegraphics[height=5.5cm]{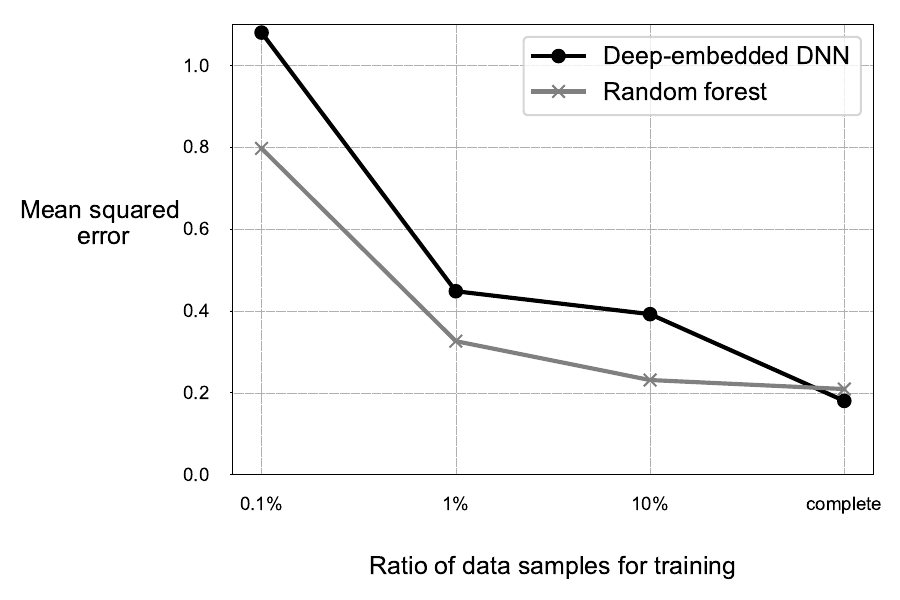}
}%
\caption{Experiment comparing number of training samples with the performance for sales forecasting. Deep neural networks still benefit from large amounts of data, whereas the performance increase of the random forest plateaus.}
\label{fig:train_samples_vs_performance}
\end{figure}

To evaluate the impact of the number of training observations on the performance, we utilize a subset of the training set and train a deep neural network and a random forest with it. \Cref{fig:train_samples_vs_performance} shows that the random forest outperforms the deep neural network up until \SI{10}{\percent} of the training observations. However, utilizing the complete dataset yields favorable results for the deep neural network. This experiment supports our statement that large datasets are needed to train deep neural networks (\cp \Cref{sec:intro})

\subsection{Proposed interpretation of deep neural networks}

The interpretation of black-box models poses a major challenge for machine learning and can significantly reduce the barriers to technology adoption. A possible remedy is given by an ex~post analysis through a visual inspection of prediction results. These same tools as for other machine learning classifiers can be applied here: for example, partial dependency plots visualize the marginal effects of one or two features having on the prediction of a machine learning model \citep{Friedman.2001}. A partial dependency plot can show whether the relationship between the target and an input feature has a certain shape, such as linear, monotonous, or more complex. In other works, Shapley values were proposed to calculate the significance of a variable by comparing what a model predicts with and without the feature \citep{Lundberg.2017}. In detail, the Shapley value of a feature value is the average change in the prediction that the model makes when the feature value is added as an input. Examples thereof are provided in the online appendix.

\begin{figure}[H]
	\centering
	\makebox[\textwidth]{%
		\includegraphics[width=.47\linewidth]{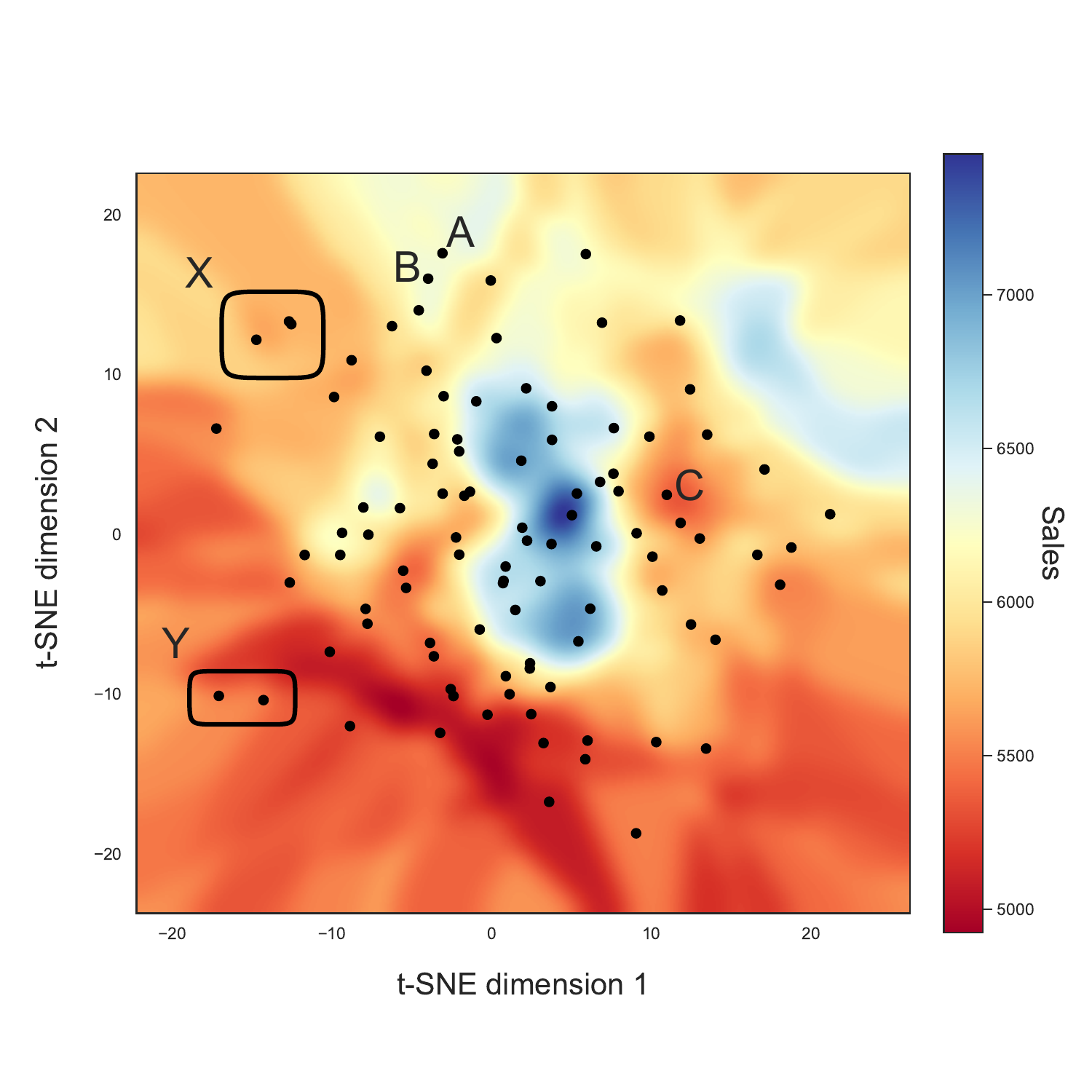}
	}%
	\caption{Proposed visual interpretation of our deep-embeded networks based on combining t-SNE dimensionality reduction with marginal effects. Black points refer to stores that were arranged based on a t-SNE dimensionality of the original categorical input (here: shown for case study~2); background color informs how embeddings are translated into an average prediction.}
	\label{fig:tsne}
\end{figure}

We propose an additional technique that is tailored to the use of embeddings as in our our deep-embedded networks. That is, we suggest to combine a lower-dimensional representation of the embeddings, specifically, the so-called t-SNE dimensionality reduction \citep{vanderMaaten.2008}, together with their marginal effects. This has obvious advantages: it allows us to visualize the embedded categorical variables, despite that the input has two or three dimensions. Here the embeddings are clustered in a way that input with similar categorical variables are located in close proximity to each other, whereas dissimilar categories are distant. This is shown by the black points in \Cref{fig:tsne}. Different from a default t-SNE plot, we additional derive marginal effects as in a partial dependence plot, so that changes in the input dimension can be studied. This yields a natural interpretation of the inner structure behind our deep-embedded network.

\vspace{-.5cm}

\Cref{fig:tsne} shows store embeddings for case study~2, which aids managerial decision-making follows. First, it helps management in identifying stores with similar characteristics but where some are over- or underperforming with respect to similar stores (see the red peaks). This should trigger a subsequent root cause whether the management, assortment, or promotion campaign needs adjustments. Second, the t-SNE plot supports management in location planning for new stores (here: indicated, \eg, by A, B, C). It displays the expected sales but also represents to what extent the new business is similar to the core operational area. Management might prefer store C over A or B despite lower expected sales since it is similar to other stores and, hence, prior experience on modi operati can be easier transferred. Third, when management wants to reduce the number of stores, this plot helps in selecting stores for disposal. Specifically, it considers trade-offs between sales and the core business area. In our example, the core business area is given by the dense inner area, whereas stores in cicles X and Y are outside of it and thus present example candidates for disposal.

\section{Discussion}
\label{sec:discussion}


Deep learning is essential in the context of big data and, for this purpose, its performance across different scenarios is compared and contrasted in this overview article. The empirical results of the different cases studies suggest that deep learning is a feasible and effective method, which can considerably and consistently outperform its traditional counterparts in both prediction and operational performance from the family of data-analytic models. As such, DNNs are able identify previously unknown, potentially useful, non-trivial, and interesting patterns more accurately than other popular predictive models such as random forest, artificial neural networks, and support vector machines. One of the reasons they yield superior results originates from the strong mathematical assumptions in traditional machine learning, whereas these are relaxed by DNNs as a result of their larger parameter space. Even though various building blocks exist, actual applications benefit from customized architectures as demonstrated by the deep-embedded architecture in this research. We obtain performance improvements between \SI{1.56}{\percent} and \SI{21.73}{\percent} across all metrics by replacing the default architecture with our proposed deep-embedded network.

\subsection{Managerial implications}


Business analytics refers to the art and science of converting (big) data into business insights for faster and more accurate managerial decision-making \citep{Chen.2012, Baesens.2016}. It is positioned at the intersection of several disciplines, of which machine learning, operations research, statistics, and information systems are of particular relevance. Accordingly, the underlying objective concerns the ability of understanding and communicating insights gleaned from descriptive, predictive, and prescriptive analytics. Business analytics is arguably the most critical enabler for businesses to survive and thrive in the stiff global marketplace, where evidence-based decisions and subsequent actions are driven from data and analytical modeling. Yet, a recent survey across \num[group-minimum-digits=3,group-separator={,}]{3000} executives reveals that 85~percent of them believe that predictive analytics implies a competitive advantage, though only one in 20 companies has adopted these techniques \citep{Ransbotham.2017}. This becomes especially crucial in the light of deep learning, which is forecasted to deepen competition over analytics. It is thus assumed that advanced analytics, as well as big data, have widespread implications for management \citep[see \eg][]{Agarwal.2014, George.2014,George.2016}.


A challenge for most managers remains with regard to how they can identify valuable use cases of predictive analytics and especially deep learning. Strategically positioned at this target, the current work presents a framework to show the viability and superiority of DNNs within the business analytics domain through several case studies. Hence, the primary message of this overview article is to review the applicability of deep learning in improving decision support across core areas of businesses operations. As a direct implication, the generic approach proposed in this work can be utilized to create an automated decision support system, which in turn would increase the quality of decisions both in terms of efficiency and effectiveness. To facilitate implementation, prior research has suggested a simple rule-of-thumb: all tasks which involve only one second of thought can be replaced by predictive analytics \citep{Ng.2016b}. This is confirmed by our experiments, since it can be safely claimed that the proposed DNN-based methodologies would bring significant financial gains for organizations across various areas of business domain as exemplified here in insurance credit scoring, IT service request predictions, and sales forecasting.


Data is the key ingredient for deep learning to become effective. In fact, a distinguishing feature of deep neural networks links to its ability to still \textquote{learn} better predictions from large-scale data as compared to traditional methods which often end up in a saturation point where larger datasets no longer improve forecasts. Hence, firms require extensive datasets and, for this topic, we refer to \citet{Corbett.2018} for an extensive discussion. 


Despite the challenges surrounding the use of deep learning, this technique also entails practical benefits over traditional machine learning. In particular, feature engineering was a critical element hitherto when deriving insights from big data \citep{Feng.2017}. Conversely, deep neural networks circumvent the need for feature engineering. This is especially helpful in the case of sequential data, such as time series or natural language \citep{Kratzwald.2018d}, where the raw data can now be directly fed into the deep neural network. A similarity becomes evident between feature engineering and embeddings, yet the latter is fully data-driven and can easily be customized to domain-specific applications \citep{Kraus.2017}.  


Deep learning can achieve higher prediction accuracies than traditional machine learning, though they are still at the embryonic stage within the areas of business analytics. Therefore, practitioners should be aware of the caveat that DNNs introduce fairly complex architectures, which, in turn, necessitate a thorough understanding and careful implementation for valid and impactful results. In addition, the value of deep learning expands beyond the scope of mere business analytics within popular areas of medicine, healthcare, engineering, and etc. with a direct societal benefit could be realized. Yet conservative estimates suggest that firms at the frontier of predictive analytics need up to 1--2 years for replicating results from research \citep{Ng.2016b}, while the actual number for the average firm is likely to be larger. Hence, practitioners can take our article as a starting point for devising a corresponding analytics strategy.  

\subsection{Limitations and possible remedies}
\label{sec:limiations_and_possible_solutions}


Further enhancing deep learning algorithmically could also be addressed by future work. We point towards three key challenges that we consider as especially relevant for the operations research community. (1)~The configuration of DNNs represents a challenging task, since it still requires extensive parameter tuning to achieve favorable results. Recently, a series of tools for automatizing the tuning process have been proposed under the umbrella term \textquote{AutoML}. This presents a promising path towards accelerating the deployment of effective DNN architectures. (2)~DNNs are currently only concerned with point estimates, while their predictions commonly lack rigorous uncertainty quantification. We hope for increasing efforts with the goal of either directly modeling distributions or even develop Bayesian variants of DNNs. (3)~Although ex post analysis have led to large steps in understanding the behavior of deep neural networks, accountability and interpretability are widely regarded as a weakness in deep learning and. Given the recent interest of policy-makers, these two active fields of research likely represent a pressing issue for a variety of applications in the near future. Potentially, estimating structural models via variational inferences \citep{Hoffman.2013,Kraus.2019b}, or interpreting attention mechanisms in neural networks could lead to more insights of the prediction \citep{Kraus.2019}.	

\subsection{Roadmap for future research}


Deep learning has great potential to create additional value for firms, organizations, and individuals in a variety of business units and domains. Yet, its actual use in the field of operations and analytics remains scarce. Hence, it is recommended that the goal of future research in the realm of business analytics could be centered around at identifying precious use cases, as well as outlining potential value gains. This further necessitates better recommendations with regard to combinations of network architectures, training routines, parameter fine-tuning that yield favorable prediction performances in the case of DNNs. For instance, further research efforts could eventually lead to novel techniques that customize network architectures when fusing different data sources. 


Similar to the other forms of machine learning methods in predictive analytics, deep learning also merely provides predictive insights, but rarely presents actual management strategies to reach the desired outcome. As a remedy to this, future studies could focus on determining how predictions can actually be translated into effective decision-making. This is another compelling direction for future research, since the OR community is more important than ever to translate predictions into decision \citep{Feng.2017}. Here we point towards inverse control \citep{Bertsimas.2014,Bertsimas.2016}, Markov decision processes and bandits as promising techniques.


There is further untapped potential as a number of innovations in the domain of deep learning have not found its way into widespread adoption. First, sequence learning as presented in this paper takes sequences as input but still yields predictions in the form of a vector with fixed dimensions. So-called \textquote{seq2seq} techniques allow to make inferences where each prediction is a sequence of arbitrarily varying length \citep{Goodfellow.2017}. This could be helpful when deep neural networks should compute routes or graphs as output. Second, datasets from practical applications often cannot warrant the necessary size that is needed for an effective use of deep learning. A remedy is given by transfer learning where a different yet related dataset is utilized for an inductive knowledge transfer. This can even facilitate an interesting strategy for domain customizations of predictive analytics \citep{Kratzwald.2019}. Third, generative adversarial networks draw upon the idea of zero-sum games and train two competing neural networks, where one provides predictions in the usual fashion, while the generates a sample input that matches a given label \citep{Goodfellow.2014}. As an illustrative example, this can essentially yield a creativity mechanism \citep{Hope.2017}, yet its benefit in OR applications still needs to be demonstrated.  

\section{Conclusion}
\label{sec:conclusion}

Business analytics refers to the ability of firms and organization to collect, manage, and analyze data from a variety of sources in order to enhance the understanding of business processes, operations, and systems. As companies generate more data at ever-faster rates, the need for advances in predictive analytics becomes a pertinent issue, which can hypothetically improve decision-making processes and decision support for businesses. Consequently, competition in terms of analytics has become prevalent as even minor improvements in prediction accuracy can bolster revenues in greater folds and thus demands a better understanding of deep learning. The case studies conducted in this overview article are purposefully selected from different areas of operations research in order to validate the fact that DNNs help in improving operational performance. However, a customized network architecture is oftentimes beneficial, such as demonstrated by our deep-embedded network which attains performance improvements between \SI{1.56}{\percent} and \SI{21.73}{\percent} across all metrics over a default, out-of-the-box architecture.

\section{Acknowledgments}
This work was part-funded by the Swiss National Science Foundation
(SNF), Digital Lives grant 10DL18-183149.



\section*{References}

\setlength{\bibsep}{0.15\baselineskip}
{\sloppy
\bibliographystyle{model5-names-no-doi}
\bibliography{literature}
}







\end{document}


\pagenumbering{gobble}%
\appendix

\section{Runtime analysis}
\Cref{tbl:runtimes} reports the runtime for optimizing the parameters inside the machine learning models.

\begin{table}[H]
	\centering
	\footnotesize
	\makebox[\textwidth]{
		\begin{tabular}{l rrr}
			\toprule
			{\textbf{Model}} & \multicolumn{3}{c}{\textbf{Optimization time (min:sec)}} \\
			\cmidrule(l){2-4}
			& {\textbf{Load forecasting}} & {\textbf{Sales forecasting}} & {\textbf{Credit scoring}} \\
			\midrule
			\multicolumn{2}{l}{\textsc{Baselines}} \\ 
			\quad ARMA/ARIMA & 0:00 & 0.07 & -- \\
			\quad Lasso & 0:00 & 0:35 & 0:07 \\   
			\quad Ridge regression & 0:00 & 0:08 & 0:04 \\  
			\quad Random forest & 1:37 & 358:10 & 129:30 \\ 
			\quad Support vector machine & 0:12 & 17:33 & 7:16 \\      
			\quad Default neural network (single-layer) &  0:07 & 0:23 & 0:46 \\
			\quad Default DNN (tuned depth) & 1:47 & 2:58 & 1:25 \\
			\midrule
			\multicolumn{2}{l}{\textsc{Proposed networks architecture}} \\
			\quad Deep-embedded DNN/LSTM & 4:17 & 7:29 & 3:24 \\
			\bottomrule
		\end{tabular}
	}
	\caption{Time (in minutes:seconds) for optimizing the machine learning models for our case studies.}
	\label{tbl:runtimes}
\end{table}

\newpage
\section{Ex~post interpretation}

\begin{figure}
	\centering
	\makebox[\textwidth]{%
		\includegraphics[width=0.5\linewidth]{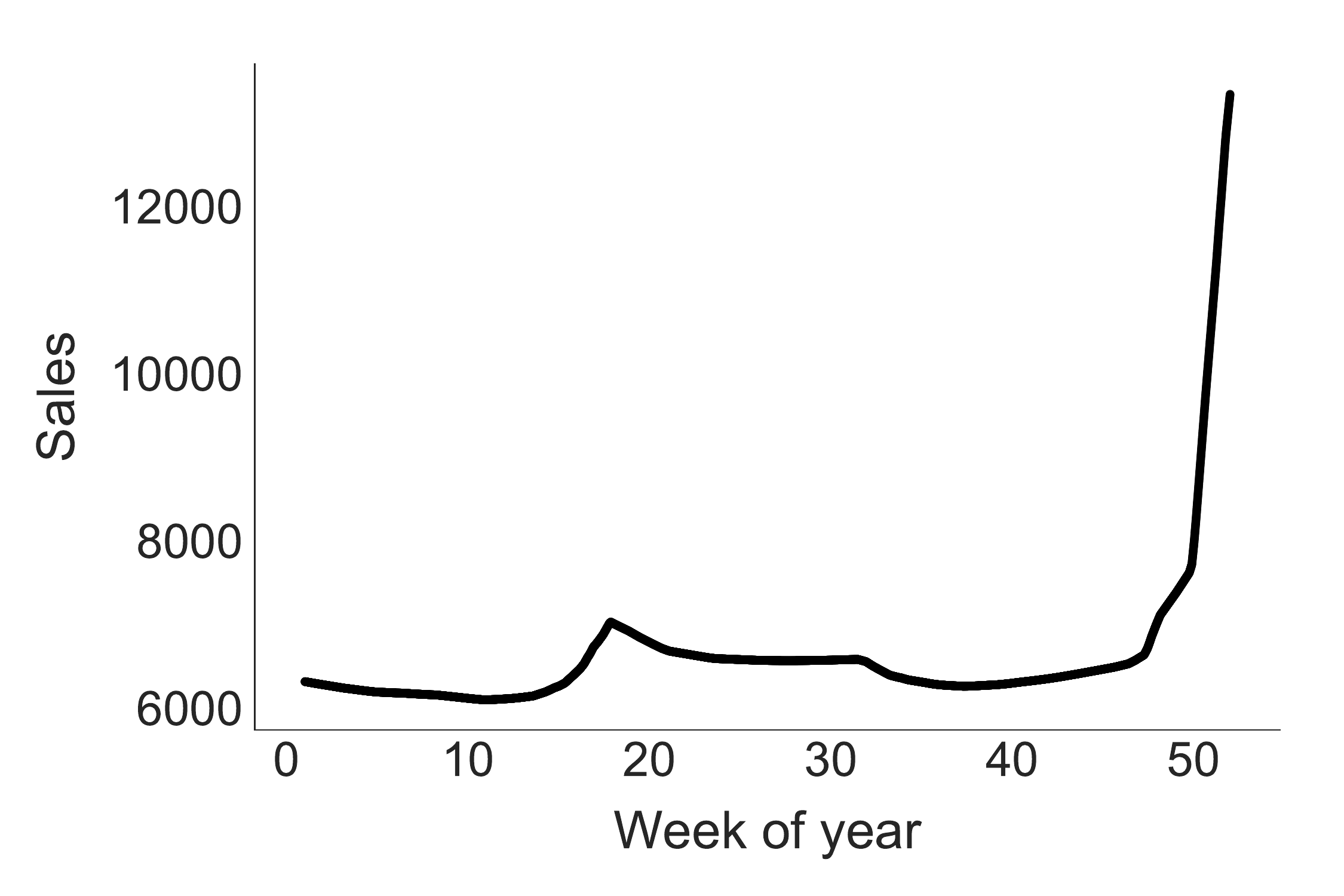}
	}%
	\caption{Partial dependence plot that shows the marginal effect of the (scaled) week-of-year variable on the prediction outcome.}
	\label{fig:pdp}
\end{figure}

\begin{figure}
	\centering
	\makebox[\textwidth]{%
	\begin{tabular}{c}
		\includegraphics[width=1.2\linewidth]{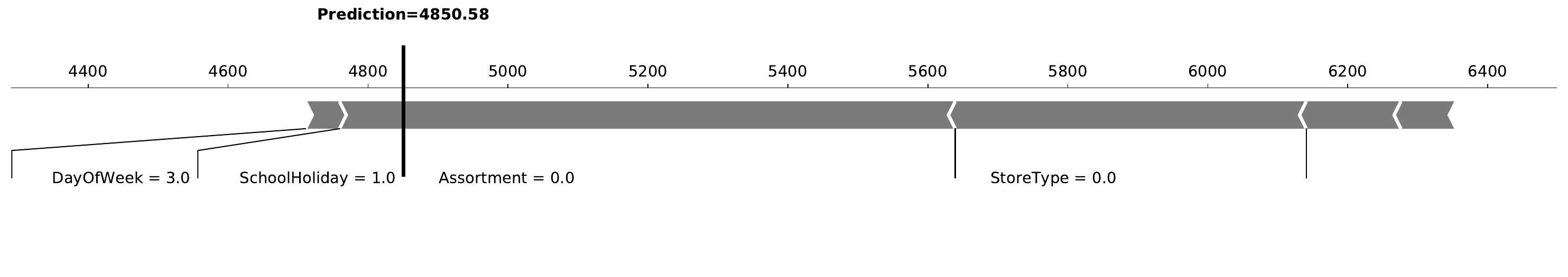} \\
		\includegraphics[width=1.2\linewidth]{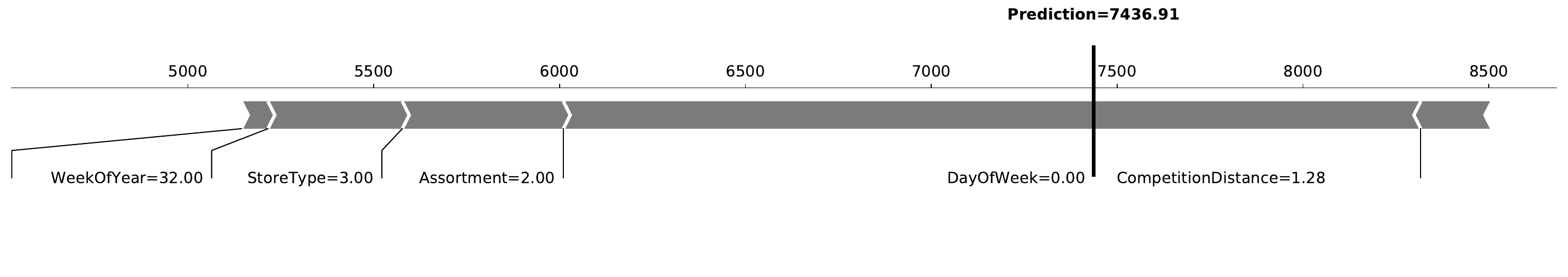}
	\end{tabular}
	}
	\caption{Shapley values show the effect of the input features on the prediction outcome (\num{4850.58} and \num{7436.91}).}
	\label{fig:shap}
\end{figure}

\Cref{fig:pdp} and \Cref{fig:shap} illustrate a partial dependence plot and Shapley values in order to interpret model behavior. The former shows increasing sales at around the 20th calendar week and, in particular, at the end of the year. \Cref{fig:shap} shows that the assortment and specific store types ($=0$) are associated with a decreasing effect on the predicted sales for the upper sample, whereas the middle of the week (day of week equals three) and school holidays lead to an increase in the predicted value.

\vspace{-.7cm}

\newpage 
\section{Hyperparameter tuning}
\Cref{tbl:grid} reports the tuning parameters used in our grid search.

\begin{table}[H]
	\centering
	\footnotesize
	\makebox[\textwidth]{
		\begin{tabular}{lll}
			\toprule
			{\textbf{Model}} & 
			{\textbf{Tuning parameter}} &
			{\textbf{Tuning range}} \\
			\midrule
			Lasso & Regularization strength $\alpha$ & $10^{-3}, \ldots, 10^{+3}$ \\
			Ridge regression & Regularization strength $\alpha$ & $10^{-3}, \ldots, 10^{+3}$ \\
			ARMA/ARIMA$^\dagger$    & Auto-regressive lags $p$ & $2,5,10,50,100,200$ \\
			& Difference iterations $d$ & $0,1,2,3$ \\
			& Moving-average terms $q$ & $0,1,2,3$ \\
			Random forest & Number of trees & $200$, $500$ \\
			& Maximum depth of trees & $2$, $5$, $10$ \\
			& Number of randomly-sampled variables & $1$, $3$, $5$, $10$ \\ 
			Support vector machine & Kernel function & Linear, radial$^\ddagger$ \\
			& Cost & $10^{-3}, \ldots, 10^{+3}$ \\
			Single-layer neural network & Learning rate & $0.001$, $0.005$, $0.01$, $0.05$ \\
			& Batch size & $32$, $64$, $256$ \\
			Default neural network & Learning rate & $0.001$, $0.005$, $0.01$, $0.05$ \\
			& Batch size & $32$, $64$, $256$ \\
			& Number of neurons in hidden layers & Task-dependent \\
			& Number of hidden layers & $1,2,3$ \\
			Deep neural network & Learning rate & $0.001$, $0.005$, $0.01$, $0.05$ \\
			& Dropout rate & $0.0$, $0.25$, $0.5$, $0.75$ \\
			& Number of neurons in hidden layers & Task-dependent \\
			& Batch size & $32$, $64$, $256$ \\
			\bottomrule
			\multicolumn{3}{l}{$^\dagger$ The ARMA and ARIMA model is only evaluated for time series prediction tasks.} \\
			\multicolumn{3}{l}{$^\ddagger$ We utilize a kernel approximation to reduce computation time.} 
		\end{tabular}
	}
	\caption{Grid search for hyperparameter tuning.}
	\label{tbl:grid}
\end{table}

\newpage
\section{Loss curves}
\begin{figure}[H]
	\centering
	\makebox[\textwidth]{%
		\begin{tabular}{cc}
			(a) Training and validation graphs of load forecasting & (b) Training and validation graphs of sales forecasting \\
			\includegraphics[width=0.5\linewidth]{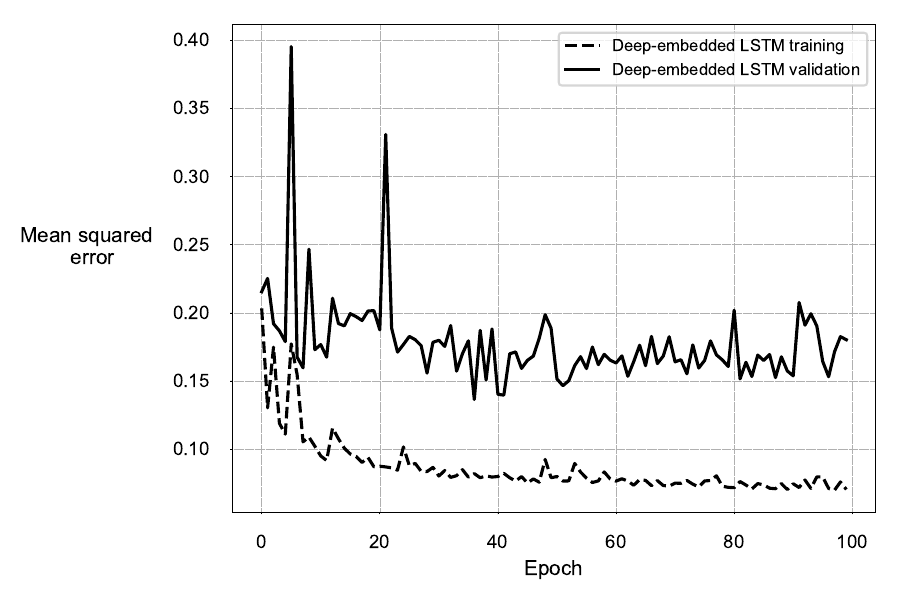} & \includegraphics[width=0.5\linewidth]{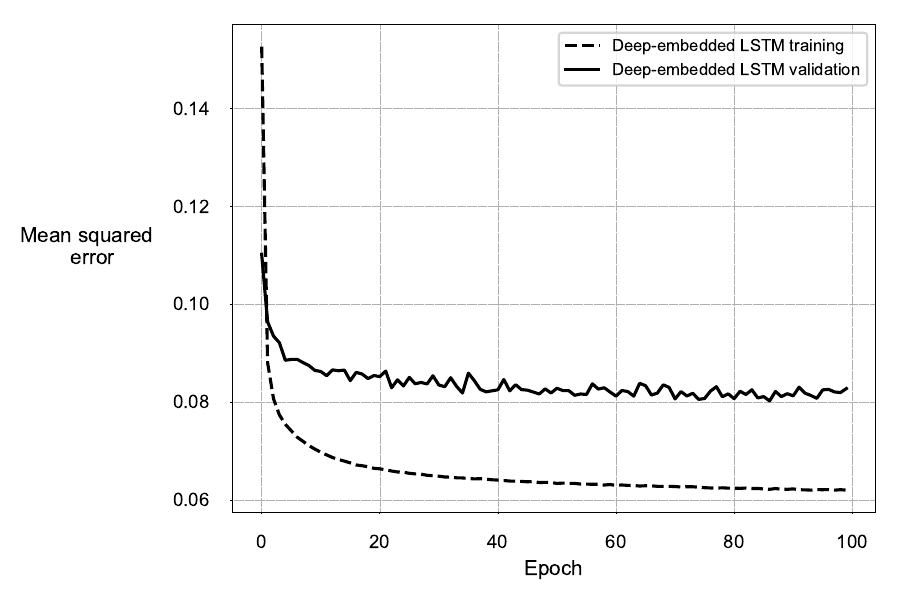} \\
		\end{tabular}
	}%
	\caption{Training and validation loss by epoch for our best performing deep neural networks on the load forecasting and sales forecasting studies. One epoch depicts a complete pass of all training samples through the optimization process. On the left, we see that the gated recurrent unit overfits after around \num{40} epochs. Therefore, additional regularization is needed to stabilize the optimization. On the right, both long short-term memory network and gated recurrent unit perform equally and converge nicely.}
	\label{fig:loss_graphs}
\end{figure}

\Cref{fig:loss_graphs} depicts our optimization processes for the LSTM and the GRU network for the load forecasting and the sales forecasting tasks. 

%





